\documentclass[10pt,twocolumn,letterpaper]{article}

\usepackage{cvpr}              

\newcommand{\ww}[1]{#1} 
\newcommand{\WW}[1]{}
\newcommand{\GR}[1]{}
\newcommand{\gr}[1]{}
\newcommand{\CO}[1]{}
\newcommand{\hx}[1]{}
\newcommand{\FZ}[1]{}

\usepackage{alphabeta}
\usepackage{pifont}
\usepackage{subcaption}
\usepackage{graphicx}
\usepackage{amsmath}
\usepackage{amssymb}
\usepackage{bm}

\DeclareMathOperator{\relu}{ReLU}

\DeclareMathOperator{\clamp}{clamp}
\DeclareMathOperator{\ceil}{ceil}
\DeclareMathOperator{\floor}{floor}

\DeclareMathOperator{\sign}{sign}

\DeclareMathOperator{\trilearp}{trilerp}

\usepackage{booktabs}
\usepackage{gensymb}
\usepackage{colortbl}
\usepackage[dvipsnames]{xcolor}
\usepackage{soul}
\usepackage{setspace}
\usepackage{float}

\definecolor{first}{rgb}{1,0.85, 0.7}
\definecolor{second}{rgb}{1,1, 0.8}
\definecolor{firstalt}{rgb}{1, 0.7, 0.7}

\newcommand{\first}{{\cellcolor{first}}}

\newcommand{\second}{{\cellcolor{second}}}

\newcommand{\name}{\alpha Surf}

\newcommand{\setLetter}{T}

\definecolor{cvprblue}{rgb}{0.21,0.49,0.74}
\usepackage[pagebackref,breaklinks,colorlinks,citecolor=cvprblue]{hyperref}

\def\paperID{125} 
\def\confName{3DV\xspace}
\def\confYear{2024\xspace}

\title{$\alpha$Surf: Implicit Surface Reconstruction for Semi-Transparent and Thin Objects with Decoupled Geometry and Opacity}

\author{Tianhao Wu\\
University of Cambridge\\
\and
Hanxue Liang\\
University of Cambridge\\
\and
Fangcheng Zhong\\
University of Cambridge\\
\and
Gernot Riegler\\
Unity\\
\and
Shimon Vainer\\
Unity\\
\and
Jiankang Deng\\
Imperial College London\\
\and
Cengiz Oztireli\\
Google Research\\
University of Cambridge\\
}

\begin{document}
\maketitle

\begin{abstract}
Implicit surface representations such as the signed distance function (SDF) have emerged as a promising approach for image-based surface reconstruction. 
However, existing optimization methods assume opaque surfaces 
and therefore cannot properly reconstruct translucent surfaces and sub-pixel thin structures, which also exhibit low opacity due to the blending effect.
While neural radiance field (NeRF) based methods can model semi-transparency and synthesize novel views with photo-realistic quality, their volumetric representation tightly couples geometry (surface occupancy) and material property (surface opacity), and therefore cannot be easily converted into surfaces without introducing artifacts.  
We present \name{}, a novel scene representation with decoupled geometry and opacity for the reconstruction of surfaces with translucent or blending effects. Ray-surface intersections on our representation can be found in closed-form via analytical solutions of cubic polynomials, avoiding Monte-Carlo sampling, and are fully differentiable by construction.
Our qualitative and quantitative evaluations show that our approach can accurately reconstruct translucent and extremely thin surfaces, achieving better reconstruction quality than state-of-the-art SDF and NeRF methods.
\end{abstract}

\section{Introduction}

Recovering object geometry as surfaces from RGB images is a long-standing problem in computer vision, with numerous practical applications such as photogrammetry, 3D asset creation, and custom 3D fabrication.
Traditional approaches rely on Structure-from-Motion~\cite{schoenberger2016sfm} and Multi-View Stereo~\cite{schoenberger2016mvs} pipelines to first reconstruct a 3D point set of the scene to which surfaces can be fitted~\cite{poisson_recon, delaunay}.
Differentiable rendering techniques have emerged as a more versatile reconstruction procedure. 
With properly derived gradients, these techniques can simultaneously learn both geometry and appearance by minimizing the error of RGB renderings.
This significantly loosens the restrictions on the choice of geometric representations to be learned.
In particular, implicit surface representations~\cite{sdfdiff, diffsdf, idr, iron}, 
\ie level sets of a scalar field such as a signed distance function (SDF), show promising results in surface reconstruction due to their robustness to complex geometry and topology.


One open challenge that has not been adequately addressed in this domain is reconstructing implicit surfaces that exhibit translucent effects. 
The prevailing assumption in earlier works is that the surface is opaque throughout, and differentiable rendering techniques for implicit surfaces only examine the intersection of rays and the nearest surface \cite{Niemeyer2020CVPR, sdfdiff, liu2020dist}. As a result, the forward rendering process in those studies cannot simulate scenes with non-opaque effects, leading to an inability to reconstruct their surfaces.

\begin{figure*}[t]
\begin{center}
  \includegraphics[width=0.95\textwidth]{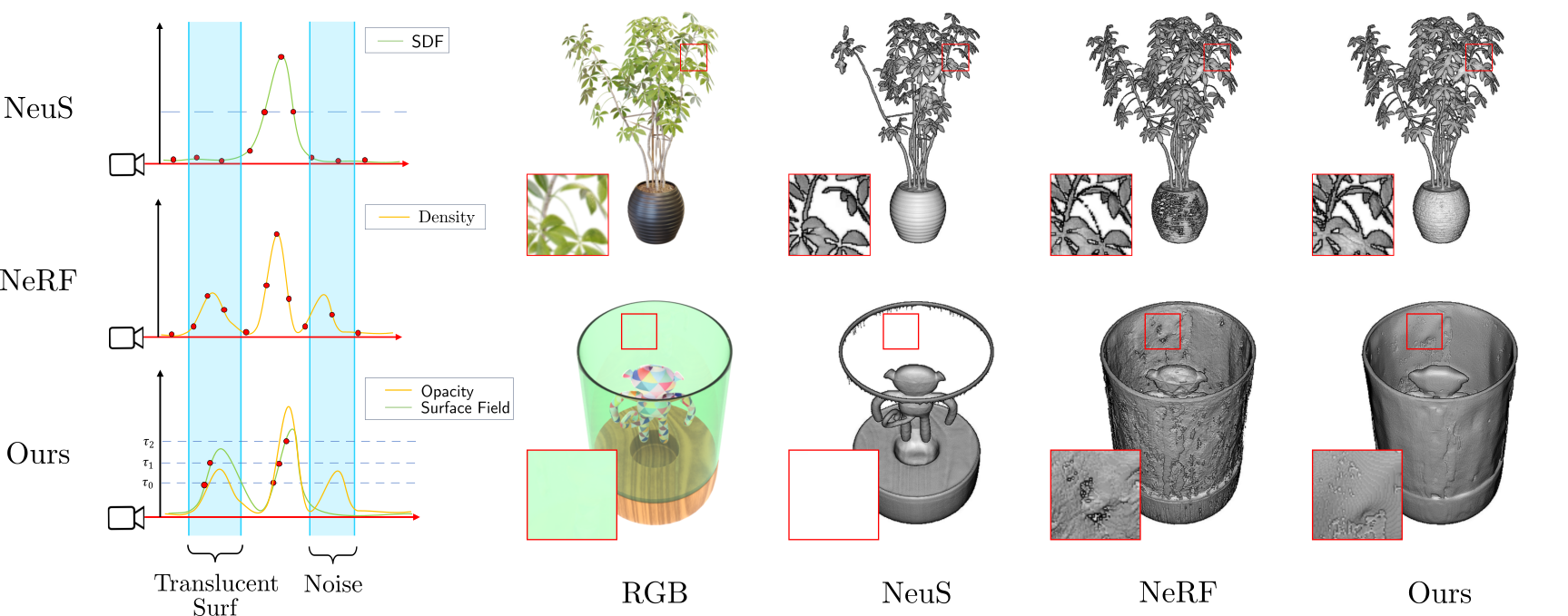}
\end{center}
    \vspace{-0.7em}
   \caption{
   \textbf{Illustration.}
   We illustrate the representations of NeuS, NeRF, and our method, as well as reconstructed surfaces. 
   NeuS~\cite{neus} uses SDF to optimize for opaque surfaces and hence misses translucent or thin surfaces with blending effects in the reconstruction.
   NeRF methods such as Plenoxels~\cite{plenoxels} can represent semi-transparency with density field, but as density couples both occupancy and opacity, surfaces extracted from it would contain holes or redundant surface floater.
   In contrast, our approach models decoupled surface and opacity fields. We use a surface field without Eikonal constraint and multiple level sets $\tau_0,\tau_1,...$ to model geometry with different levels of confidence and opacity, and utilize a closed-form intersection formula to enable differentiable rendering, and hence can accurately reconstruct surfaces exhibiting semi-transparency. 
   }
\label{fig:teaser}
\vspace{-10pt}
\end{figure*}

\begin{figure}[t]
\begin{center}
  \includegraphics[width=0.4\textwidth]{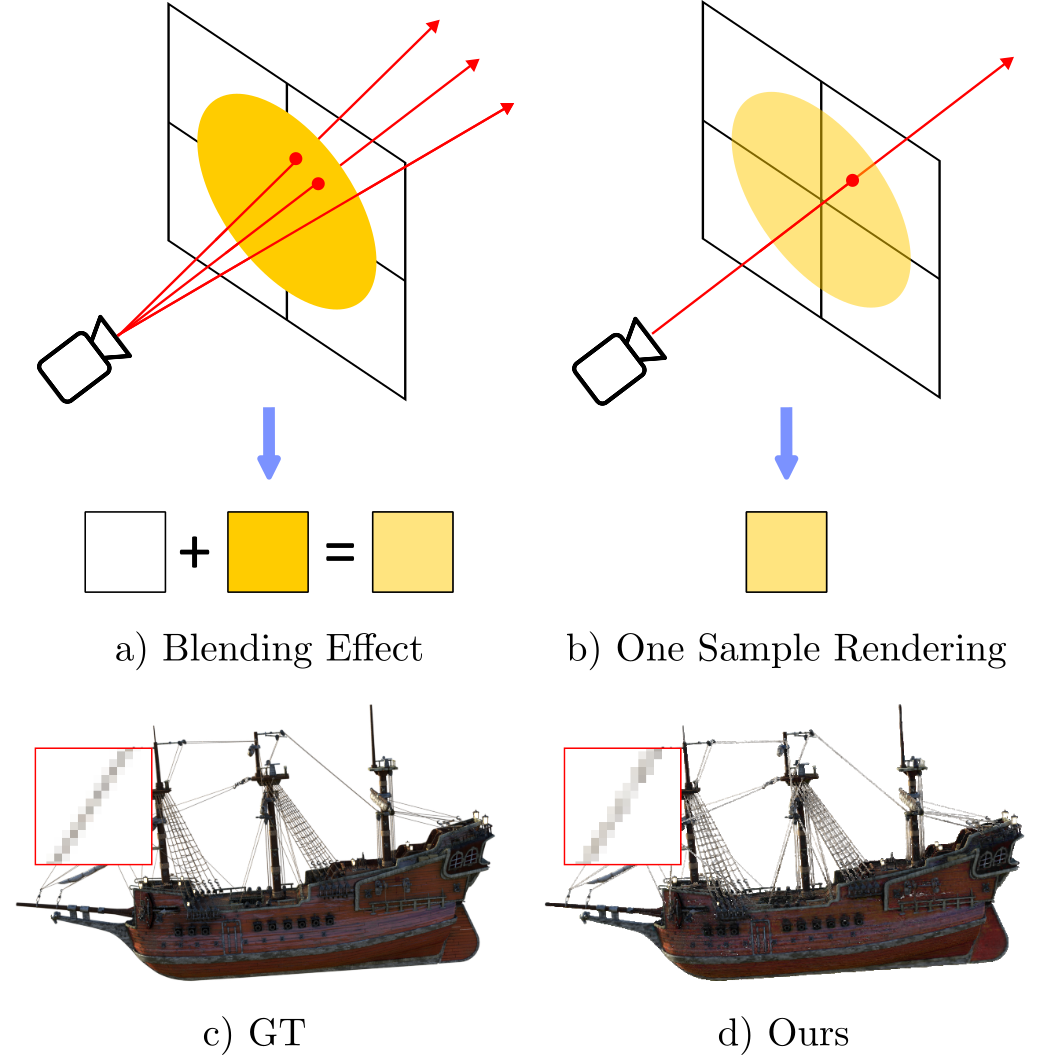}
\end{center}
   \caption{\textbf{Blending effect.} a) Real-world capture takes incoming light from multiple rays per pixel. Pixels that are partially occupied by an opaque object are therefore rendered as a mixture of the object and background color. b) With one-sample rendering in reconstruction, it becomes necessary for extremely tiny objects to be modeled as semi-transparent for the pixel color to match the ground truth. c), 
   d) Our representation is fully capable of representing this phenomenon, leading to the accurate surface reconstruction of thin structures.
   }
\label{fig:blending}
\vspace{-15pt}
\end{figure}

Modeling opaqueness is not solely crucial for reconstructing translucent surfaces, but is also necessary for the recovery of extremely thin surfaces with sub-pixel silhouette. As illustrated in Figure~\ref{fig:blending}, when rendering a thin structure that only partially occupies a pixel with an insufficient number of sampled rays, opaque foreground objects must be treated as semi-transparent in order to achieve an accurate pixel color blending the foreground and background. 
Methods that neglect this feature may fail to capture the correct reconstruction of thin structures; see Figure~\ref{fig:teaser}.

Differentiable volume rendering techniques ~\cite{nerf} have demonstrated considerable success in rendering translucent and thin objects by integrating radiance along a ray through an entire scene.
Their geometric representation is a density field where a surface can be implicitly defined by applying a density threshold.
However, density represents both occupancy (geometry) and transparency (material) in a tightly coupled manner, posing a non-trivial challenge in choosing the threshold for surface extraction. 
A high threshold may exclude translucent surfaces from the reconstruction, whereas a low threshold results in the erroneous reconstruction of density floaters and redundant surfaces that are not part of the desired reconstruction.
NeuS~\cite{neus} attempts to alleviate this issue by combining SDF with volume rendering, but it still assumes an opaque surface at the end of optimization and hence fails to reconstruct surfaces with semi-transparency.
Figure~\ref{fig:teaser} illustrates failure cases of both types of methods on translucent or thin surfaces. 
\ww{Recent Gaussian Splatting methods~\cite{gaussian_splatting, sugar} uses anisotropic 3D Gaussians to achieve efficient rendering, but they similarly ignore the occupancy-transparency ambiguity and only retain opaque Gaussians for surface extraction, and hence cannot reconstruct thin or translucent surfaces.}

Two key requirements must be fulfilled for faithfully reconstructing surfaces of translucent and thin objects: 1) a representation that explicitly decouples geometry and materials; and 
2) a differentiable rendering pipeline that considers more than the nearest ray-surface intersection, while also enabling gradient flows to both surface and opacity.
To this end, we introduce \name{}, a novel surface representation based on a grid structure without neural networks. 
We use separate values on the grid to represent geometry, opacity, and appearance.
We define the surface as multiple level sets of a continuous scalar field, where the field itself is given by a trilinear interpolation of the grid values. Unlike previous methods, our representation does not require the scalar field to be an SDF subject to the Eikonal constraint.
An important property 
of our approach is that the \emph{exact} intersection points between a ray and all the surfaces, regardless of whether they are opaque or transparent, can be determined by analytically solving a cubic polynomial. The \emph{closed-form} solution allows for full \emph{differentiability} to both geometry and material in our forward rendering process, which simulates the semi-transparency effects via alpha compositing of intersection points. 

We further propose a series of initialization and optimization strategies that are designed to facilitate efficient and accurate reconstruction. 
The total training time of our method is around 30 minutes.
We evaluate our method on an extended version of the NeRF synthetic dataset~\cite{nerf}, which contains 8 original scenes and 16 additional objects with challenging thin structures or translucent materials. We show that our method is capable of reconstructing surfaces with better quality than the existing SDF and NeRF based methods.

In summary, our contributions are: 
1) A novel grid-based scene representation for implicit surface reconstruction, specialized for translucent and thin objects. It incorporates a closed-form and differentiable evaluation of all surface intersections along the ray, and properly decouples surface geometry and opacity.
2) Initialization scheme utilizing fast Plenoxels \cite{plenoxels} training and optimization with truncated volume rendering and surface smoothness constraint for efficient training and accurate reconstruction.
3) We show superior reconstruction quality compared to state-of-the-art methods on synthetic and real-world scenes featuring thin and translucent objects.

\section{Related Works}

\paragraph{Neural Radiance Fields} 
NeRF~\cite{nerf} is a plenoptic function of volume density and view-dependent appearance. Its differentiable volume rendering allows robust image-based 3D reconstruction and motivated many works in high-quality novel view synthesis~\cite{mipnerf360, refnerf, hdrnerf, nerf++}, 3D asset synthesis and editing ~\cite{gram, graf, dreamfusion, instructnerf2nerf}, few-shot reconstruction \cite{pixelnerf, zero123, SinNeRF, LEAP}, and efficient rendering~\cite{instantNGP, kilonerf, plenoxels, directvoxgo, plenoctrees, gaussian_splatting}. Despite the outstanding novel view synthesis performance, their geometry tends to produce artifacts such as sparse density floaters and inner volume~\cite{mipnerf360, refnerf, nerf++}. 
Direct surface extraction on the density field hence suffers from those artifacts, whereas the depth extraction method does not guarantee complete surfaces and requires additional surface reconstruction \cite{tsdf, alpha_shape, poisson_recon}.

\vspace{-5pt}
\paragraph{Signed Distance Field} 
SDF has been extensively employed with differentiable rendering methods to reconstruct surfaces from multi-view images. 
IDR~\cite{idr} proposes a differentiable sphere-tracing algorithm and optimizes the surface together with a volumetric BRDF. 
VolSDF~\cite{volsdf} maps SDF to volume density via Laplacian CDF and optimizes the SDF via volume rendering.
NeuS~\cite{neus} similarly maps an SDF to unbiased weights in the volume rendering equation via a logistic sigmoid function. 
NeuS has motivated several further applications in different areas, such as sparse view surface reconstruction~\cite{sparse_neus, volrecon}, fast reconstruction~\cite{neus2, voxel_surf, hash_sdf}, and finer details modeling \cite{HFS, neuda, HFS, neuralangelo, GeoNeus}. 
A crucial and common limitation in existing SDF optimization methods is the assumption of surface opaqueness. Even the methods that utilize volume rendering in surface reconstruction still enforce convergence to opaque surfaces. Hence, they cannot properly reconstruct semi-transparent surfaces and suffer from thin structures with strong blending effects. 

\vspace{-5pt}
\paragraph{Transparent Object Reconstruction} 
\ww{Many works have tried to address the problem of transparent object reconstruction~\cite{neto, laser_transparent_recon, through_glass}. They focus on thick and fully transparent objects, and aim to reconstruct the shape by solving the complicated light transportation within the objects. Most methods require additional supervision such as environment matting, which can only be obtained with checkerboard-patterned backgrounds. In comparison, we reconstruct thin translucent surfaces from RGB images alone with no other supervision. NeRRF~\cite{chen2023nerrf} is a method that similarly reconstructs surfaces from RGB images and mask supervisions.}


\section{Method}

\begin{figure*}[t]
\begin{center}
  \includegraphics[width=\textwidth]{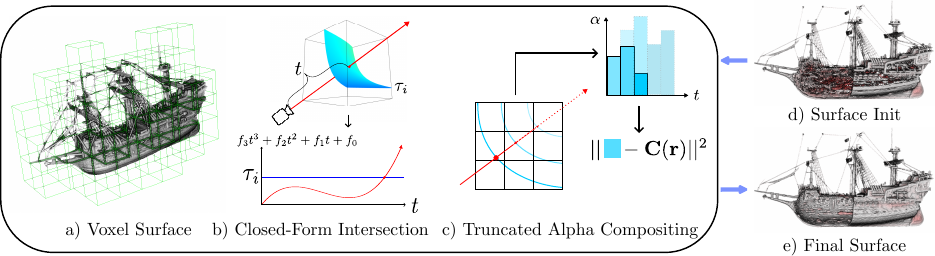}
\end{center}
   \caption{\textbf{Method.} 
   a) Our surface representation is based on a voxel grid storing explicit values, without neural networks; see Section~\ref{sec:representation}. 
   b) We utilize a closed-form and differentiable method to compute ray-surface intersection. This is achieved by solving a cubic polynomial of depth $t$ with known parameters $f_0,...,f_3$ and $\tau_i$; see Section~\ref{sec:rendering}. 
   c) We incorporate surface-specific regularization such as truncated alpha compositing to obtain clean and accurate surfaces; see Section~\ref{sec:optimization}.
   d) We utilize a coarse initialization via Plenoxels~\cite{plenoxels} to start with roughly correct yet noisy surfaces. 
   e) The optimization results in clean and complete surfaces in the end.}
\label{fig:method}
\end{figure*}


Given a set of multi-view posed RGB images, our goal is to recover an implicit surface of the scene objects, particularly in cases involving translucent or thin surfaces.
%
Towards this, we propose \name{}, a grid-based representation where the grid contains values pertaining 
to the surface's geometry and material properties (Figure~\ref{fig:method}a). 
This enables a closed-form evaluation of all the ray-surface intersections (Figure~\ref{fig:method}b), and thus a fully differentiable alpha composition (Figure~\ref{fig:method}c) to render the intersection points.
Our method pre-trains Plenoxels~\cite{plenoxels} to efficiently initialize a coarse surface (Figure~\ref{fig:method}d), and through the optimization of a photometric loss and additional regularizations, we are able to reconstruct clean and accurate surfaces (Figure~\ref{fig:method}e).

\subsection{Representation}\label{sec:representation}

\noindent \textbf{Surface} \hspace{0.5em}
Following voxel-based representations~\cite{plenoxels, voxel_surf}, we represent the surface as the level sets of a continuous scalar field $\delta: \mathbb{R}^3 \rightarrow \mathbb{R}$.
For each spatial coordinate $\mathbf{x}$ within a voxel $v_\mathbf{x}$, the value of the scalar field $\delta(\mathbf{x})$ is obtained by 
the trilinear interpolation of scalars stored at the voxel vertices:
\begin{equation}
    \delta(\mathbf{x}) = \trilearp(\mathbf{x}, \{\hat{\delta}_i\}_{i=1}^{8}) 
    \label{eq:trilearp_surf}
\end{equation}
\noindent
where 
$\{\hat{\delta}_i\}_{i=1}^{8}$ are the surface scalars stored at its eight adjacent vertices.
The surface is then implicitly defined as level sets on the scalar field. Specifically, given a set of $n$ constants $\bm{\tau} = \{\tau_i\}_{i=1}^n$ which we refer to as \emph{level values}, the surface $\mathcal{S}$ is defined as:
\begin{equation}
\mathcal{S} = \{\mathbf{x} \in \mathbb{R}^3 | \exists \tau \in \bm{\tau}: \delta(\mathbf{x}) = \tau\} \,.
\end{equation}
The cardinality $n$ and values of $\bm{\tau}$ are determined through hyperparameter tuning. 
Note that this implicit surface field does not model distance to the closest surface as in SDF, therefore, it is not subject to the Eikonal constraint.
The motivation behind the multi-level set is tightly related to our NeRF initialization strategy and is further explained in Sec~\ref{sec:optimization}.


\noindent \textbf{Opacity and Appearance} \hspace{0.5em}
Within the same voxel grid, we also model the surface opacity denoted as $\alpha(\mathbf{x})$ and view-dependent appearance denoted as $\mathbf{c}(\mathbf{x}, \mathbf{d})$ 
in the same explicit style as in Plenoxels~\cite{plenoxels}, which represents $\mathbf{c}(\mathbf{x}, \mathbf{d})$ as coefficients of the spherical harmonic (SH) function that maps view direction $\mathbf{d}$ to radiance. Trilinear interpolation is applied to obtain opacity and SH coefficients at surface locations. 
Note that although $\alpha(\mathbf{x})$ is essentially defined across all valid voxels in the 3D space, it only represents surface material property rather than geometry, and hence is only meaningful where surface exists.

%
%
%
\subsection{Differentiable rendering}
\label{sec:rendering}

A key feature in the rendering process of our representation is that it does not involve any Monte-Carlo sampling as in NeRF or sphere tracing as in SDF-based methods. Instead, it relies on ray-voxel traversal and directly takes samples at the ray-surface intersections found through a \textit{closed-form} and fully \textit{differentiable} function. Specifically, for each camera ray with origin $\mathbf{o}$ and direction $\mathbf{d}$, we first determine the set of voxels it traverses through and substitute the ray equation $\mathbf{r}(t) = \mathbf{x} = \mathbf{o} + t \mathbf{d}$ into Eq\onedot~\ref{eq:trilearp_surf}.
for each voxel $v_\mathbf{x}$. 
By setting $\delta(\mathbf{x})$ to each level set value $\tau_i$, we obtain a cubic polynomial $\tau = f_3 t^3 + f_2 t^2 + f_1 t + f_0$ with a single unknown $t$.
%
%
%
$f_0, \ldots, f_3$ are known coefficients obtained from camera origin $\mathbf{o}$, ray direction $\mathbf{d}$ and voxel surface scalars $\{\hat{\delta}\}_{i=1}^{8}$. 
We refer to the supplementary materials 
for a detailed derivation. The real roots of this cubic polynomial can then be found in closed-form via Vieta's approach~\cite{numerical_recipes}, which minimizes the error caused by the numerical precision issues. Roots that give intersections outside of each corresponding voxel $v_\mathbf{x}$ are deemed invalid and removed. The remaining intersections are ordered from near to far and taken as samples for rendering. For all intersection points $\{t_i\}$ along the ray, we obtain their surface opacity $\alpha$ and view-dependent radiance $\mathbf{c}$ through trilinear interpolation and evaluating the SH function, and then perform alpha compositing to render the pixel color:
%
\begin{align}
\hat{C}(\mathbf{r}) &= \sum_{i=1}^N T_\alpha(t_i) \alpha(t_i) \mathbf{c}(t_i)
\label{eq:alpha_compositing}
\\
T_\alpha(t_i) &= \prod_{j=1}^{i-1} (1-\alpha(t_j))
\label{eq:alpha_transmittance}
\end{align}
where we simplify our notation as $\alpha(t_i)\equiv \alpha(\mathbf{r}(t_i))$ and $\mathbf{c}(t_i)\equiv \mathbf{c}(\mathbf{r}(t_i), \mathbf{d})$.
As the samples are taken analytically through cubic polynomial root solving, which is a fully differentiable function, we naturally have gradients from the photometric loss back to the implicit surface field, without requiring any approximation or re-parameterization trick. 

\subsection{Optimization}\label{sec:optimization}

\paragraph{NeRF Initialization} 
One advantage of our representation is that we can easily initialize coarse surfaces from a pre-trained NeRF
. In practice, we use Plenoxels~\cite{plenoxels}, a grid-based NeRF method that can be trained efficiently within 10 minutes. 
After fitting a NeRF, we obtain a density field $\sigma(\mathbf{x})$ from which we select a set of raw level values $\bm{\tau}_\sigma$ to define the initial surfaces. 
The values of $\bm{\tau}_\sigma$ are selected by first determining an upper bound of the initialized density field, then evenly dividing the range by $n$ level values. The number of level sets $n$ is determined by hyperparameter-tuning.
We then normalize the density $\sigma(\mathbf{x})$ to be used as our initial surface scalars $\delta(\mathbf{x})$, and normalize the raw level values to be used as our level values:
\begin{align}
    \delta(\mathbf{x}) &= \frac{\sigma(\mathbf{x}) - \tilde{\tau}_\sigma}{\overline{||\nabla \sigma||}}
    \text{, }
    \bm{\tau} = \left\{\frac{\tau_\sigma - \tilde{\tau}_\sigma}{\overline{||\nabla \sigma||}} \mid \tau_\sigma \in \bm{\tau}_\sigma \right\}
\end{align}
\noindent
where $\tilde{\tau}_\sigma$ is the median of the chosen raw level values,
$\overline{||\nabla \sigma||}$ is the average over the norms of the finite difference gradient of the voxelated density field. $\overline{||\nabla \sigma||}$ is used to keep the initialized surface field within a constant range to make optimization easier. 
Note that $\bm{\tau}_\sigma$ is a hyperparameter and is defined only to make the selection of initial surface level values more convenient, whereas $\bm{\tau}$ is the actual level set values of the implicit surface field used throughout the optimization.

Although in theory, a single level set is sufficient to represent the surfaces, our multi-level set initialization scheme allows the geometric information from NeRF to be maximally preserved and inherited to the surface field --
In NeRF, high density regions represent opaque geometries with high confidence, whereas low density regions 
can be either low-confidence surfaces or translucent surfaces and hence are ambiguous. 
Multi-level sets initialization therefore simultaneously captures high-confidence opaque geometry with higher level values, and translucent geometry or low-confidence noise region with lower level values. 
Later optimization can then easily identify and remove those redundant noise surfaces. 
The effect of multi-level set initialization is also shown empirically in Sec~\ref{sec:abla}.


We also initialize the opacity and SH field from the pre-trained Plenoxels to facilitate optimization. After taking the raw density values $\sigma$ in the voxel grid, they are rescaled with a constant $s_\sigma$ to be used as raw surface opacity $\sigma_\alpha$, then mapped to opacity through a combined exponential-ReLU activation:
%
%
%
\begin{align}
\sigma_\alpha &= s_\sigma \ \sigma
\text{, }
\alpha = 1 - \exp\left(-\relu(\sigma_\alpha)\right) 
\end{align}
Note that this activation resembles the mapping from discretized sample density to opacity in volume rendering where the step size term is replaced by the rescaling factor $s_\sigma$~\cite{nerf}. Unlike sigmoid which has a vanishing gradient towards 0, this activation can more easily
encourage sparsity in the surface opacity to remove redundant surfaces. The rescaling $s_\sigma$ is to ensure that initialized raw opacities do not map to $\alpha$ with too high values after removing the step size term, causing the gradients to be saturated. 
Note that, during training, we update $\sigma_\alpha$ as the training parameters rather than $\alpha$.
The SH coefficients are also initialized from the pre-trained Plenoxels and further optimized.

In comparison to our approach, SDF methods such as NeuS~\cite{neus} cannot easily take the advantages of initializing from Plenoxels or other NeRF-based methods 
due to two key aspects: 1) initializing the weights of the network to become an SDF that matches the coarse shape learned by NeRF is difficult, as it requires additional optimization to fit the shape while satisfying the Eikonal constraint;
 and 2) even for explicit grid-based SDF methods~\cite{sdfdiff, diffsdf}, algorithms such as fast sweeping~\cite{fast_sweep2, fast_sweep} are required to explicitly assign the grid values as distance to closest surface~\cite{sdf_redistance, sdf_redistance2, sdf_redistance3}. 
This process can be done efficiently, but it no longer preserves the information in the initialized density field, making the removal of redundant surfaces and recovery of missing surfaces more difficult. 


\paragraph{Truncated Alpha Compositing} 
A critical artifact in NeRF methods, including Plenoxels, is that they tend to learn low-density surfaces and inner volumes that are only visible from certain angles to represent high-frequency view-dependent appearance~\cite{refnerf}. This leads to very noisy inner surfaces when initialized from Plenoxels; see Figure~\ref{fig:method}. 
To regularize those artifacts, we first remove ray intersections on backward-facing surfaces, and define the remaining set $\mathcal{\setLetter}$ of intersection points to be considered for rendering and regularization: 
\begin{equation}
    t_i \in \mathcal{\setLetter}\,\,\text{if}\, \underbrace{\mathbf{n}(t_i) \cdot \mathbf{d} < 0}_{\text{back-face culling}}
\end{equation}
\noindent
where $t_i$ is an original intersection, $\mathbf{n}(t_i) = \frac{-\nabla \delta(t_i)}{||\nabla \delta(t_i)||}$ is the surface normal, and 
$\mathbf{n}(t_i) \cdot \mathbf{d}$ checks whether the surface is facing backwards. This is similar to back-face culling in rendering. We then apply additional constraints by incorporating a truncated version of alpha compositing, where the later intersections along the ray are down-weighted with a truncated Hann window~\cite{nerfies} to give less contribution to the rendering. Specifically, we obtain the re-weighted sample opacity as:
\begin{align}
    \alpha^*(t_i) &= \gamma(i-1) \ \alpha(t_i) 
    \label{eq:truncated_alpha}
    \\
    \gamma(x) &= \left(1-\cos(\pi \clamp(a-x, 0, 1))\right) / 2
\end{align}
\noindent
where $i$ is the index of the intersection starting from 1. Note that this is the index with the back-face intersections excluded. 
During training, we linearly reduce $a$ so that $\gamma(i-1)$ is only greater than zero for a smaller range of $i$.
I.e., only the first $\ceil(a)$ intersections are kept and the rest are truncated. 
We now re-define our alpha compositing using this truncated version of opacity values: $\hat{C}^*(\mathbf{r}) = \sum_{t_i \in \mathcal{\setLetter}} T_\alpha^*(t_i) \alpha^*(t_i) \mathbf{c}(t_i)$, where $ T_\alpha^*(t_i) = \prod_{j=1}^{i-1} (1-\alpha^*(t_j))$.
\paragraph{Regularization} 
We additionally apply regularizations to enforce smooth surfaces and mitigate the artifacts inherited from initialization. 
As previously mentioned, we initialize with multiple level sets to preserve geometry priors. However, this can lead to redundant surfaces and potentially deteriorate the reconstruction quality. 
We hence apply a surface convergence loss to each 
ray-surface intersection with non-trivial truncated opacity to converge different level surfaces together: 
%
%
\begin{equation}
\begin{split}
    \mathcal{L}_c(\mathbf{r}) &= \sum_{t_i \in \mathcal{\setLetter}} |\tilde{t} - t_i| \ \mathbb{I}[\alpha^*(t_i) > 10^{-8}] 
\end{split}
\end{equation}
%
\noindent
where $\tilde{t}$ is the depth of the sample with the highest rendering weight $w(t_i) = T^*_{\alpha}(t_i) \alpha^*(t_i)$ on the ray, and $\mathbb{I}$ is the indicator function. This loss remedies the out-growing surfaces initialized from multi-level sets by encouraging them to move towards the actual surface location. 
The surface is also smoothed via a combination of an L1 and squared L2 
normal smoothness loss and a total variation (TV) loss applied on the surface field:
%
\begin{align}
\mathcal{L}_{\mathbf{n}_1} &= \frac{1}{|\mathcal{V}|}  \sum_{\mathbf{x} \in \mathcal{V}} | \nabla \mathbf{n}(\mathbf{x}) |
\label{eq:l1_norm_reg}
\text{, }
\mathcal{L}_{\mathbf{n}_2} = \frac{1}{|\mathcal{V}|}  \sum_{\mathbf{x} \in \mathcal{V}} || \nabla \mathbf{n}(\mathbf{x}) ||^2
\\
\mathcal{L}_\delta &= \frac{1}{|\mathcal{V}|}  \sum_{\mathbf{x} \in \mathcal{V}} ||\nabla \hat{\delta}(\mathbf{x})||
\end{align}
\noindent
where $\mathcal{V}$ contains the spatial coordinates of all voxels,
$\nabla \hat{\delta}(\mathbf{x})$ is the surface gradient at voxel grids obtained using finite differences on the adjacent grid values. Note that normal regularization is applied regardless of whether the voxel contains a valid surface or not, as it can help to smooth the overall surface field instead of just the actual surface. While the normal regularization $\mathcal{L}_{\mathbf{n}_1}, \mathcal{L}_{\mathbf{n}_2}$ encourages smooth surfaces in a local scope, it alone struggles to remove redundant inner surfaces. $\mathcal{L}_\delta$ is more effective for regularizing redundant surfaces with large errors; 
see Supplementary.
Note that the use of $\mathcal{L}_\delta$ is only possible because our implicit surface field is not an SDF and does not need to satisfy the Eikonal constraint.

Finally, we regularize the opacity field with a weight-based entropy loss adapted from~\cite{infonerf} on each ray and an L1 sparsity loss:
\begin{align}
\mathcal{L}_\mathcal{H}(\mathbf{r}) &= - \sum_{t_i \in \mathcal{\setLetter}} \bar{w}(t_i) \log(\bar{w}(t_i))
\label{eq:w_entropy}
\\
\text{where } \bar{w}(t_i) &= \frac{T^*_{\alpha}(t_i) \alpha^*(t_i)}{\sum_{t_j \in \mathcal{\setLetter}}T^*_{\alpha}(t_j) \alpha^*(t_j)}
\\
\mathcal{L}_\alpha &= \frac{1}{|\mathcal{V}'|} \sum_{\mathbf{x} \in \mathcal{V}'} |\relu(\sigma_\alpha(\mathbf{x}))|
\label{eq:alpha_sparse} 
\end{align}
\noindent
where 
$\mathcal{V}'$ is 10\% of all existing voxels sampled uniformly at each iteration. They together encourage surfaces to have more concentrated and minimal opacity. Note that $\mathcal{L}_\mathcal{H}(\mathbf{r})$ does not always give beneficial regularization for scenes with semi-transparent materials, but we empirically found that having this term with a small weight can help remove the noise in the surface opacity.

Given a batch $B$ of rays,  the optimization target is:
\begin{align}
\mathcal{L} &= \frac{1}{|B|} \sum_{\mathbf{r} \in B} \Big( || C(\mathbf{r}) - \hat{C}^*(\mathbf{r})||^2 
+ \lambda_c \mathcal{L}_c(\mathbf{r}) 
\nonumber 
+ \lambda_\mathbf{n_1} \mathcal{L}_\mathbf{n_1} 
\\
&+ \lambda_\mathbf{n_2} \mathcal{L}_\mathbf{n_2} 
+ \lambda_\delta \mathcal{L}_\delta 
+ \lambda_\mathcal{H} \mathcal{L}_\mathcal{H}(\mathbf{r})
+ \lambda_\alpha \mathcal{L}_\alpha \Big) 
\end{align}
\noindent
where $\lambda_c, \lambda_\mathbf{n_1}, \lambda_\mathbf{n_2}, \lambda_\delta, \lambda_\mathcal{H}, \lambda_\alpha$ are hyperparameters. 
The optimization targets include surface scalar $\delta$, grid SH coefficients and raw opacity $\sigma_\alpha$.

\section{Evaluation}

\begin{figure}[t]
  \centering
\hfill
  \begin{minipage}[]{0.45\textwidth}
\small
\centering
\singlespacing
\begin{tabular}{lccc}
\toprule
{} &   Thin & Translucent &  Avg \\
\midrule
Plen   &          0.526 &          0.761 &          0.644 \\
Mip360 &          1.445 &          3.063 &          2.254 \\
NeuS          &          1.048 &          2.344 &          1.696 \\
HFS           &          0.925 &          3.698 &          2.312 \\
neuralangelo  &          0.424 &          1.125 &          0.774 \\
NeRRF         &          2.349 &          2.086 &          2.218 \\
Ours          &  \first{0.284} &  \first{0.624} &  \first{0.454} \\
\bottomrule
\end{tabular}
    \captionof{table}{\textbf{Chamfer distance $\downarrow \times 10^{-2}$ on synthetic datasets.} We highlight the \colorbox{first}{best} methods. 
    We justify the choice of density level set values for Plenoxels and MipNeRF360 in the supplementary. 
    } 
    \label{tab:chamfer}
  \end{minipage}
  \hfill
  \begin{minipage}[]{0.47\textwidth}
  \includegraphics[width=\textwidth]{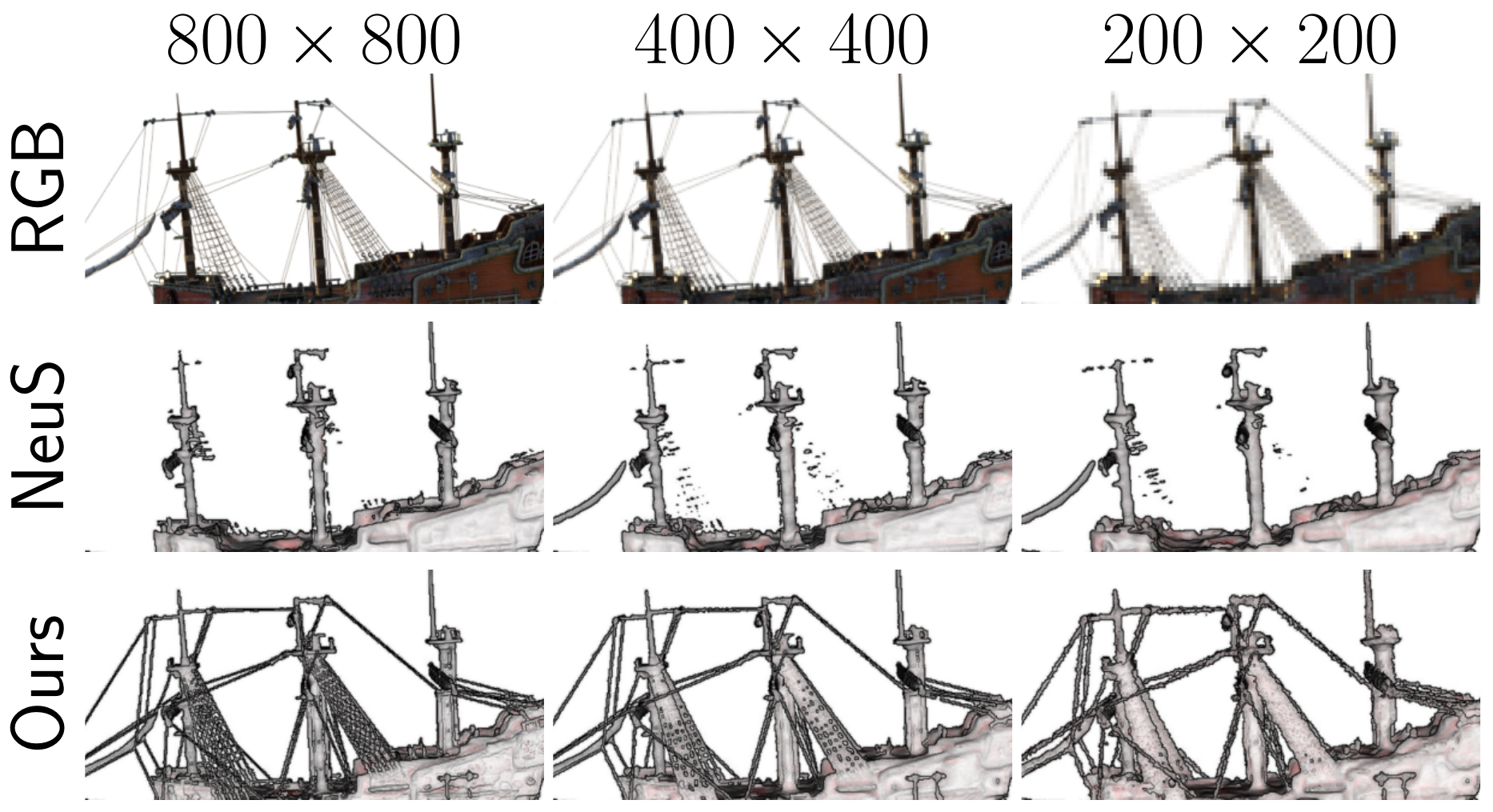}
   \caption{\textbf{Surface reconstruction with lower resolutions.} Our method can reconstruct thin surfaces under a low resolution, where thin structures such as ropes heavily blend with the background.
   }
\label{fig:low_res}
  \end{minipage}
\end{figure}

\begin{figure*}[t]
\begin{center}
  \includegraphics[width=0.95\textwidth]{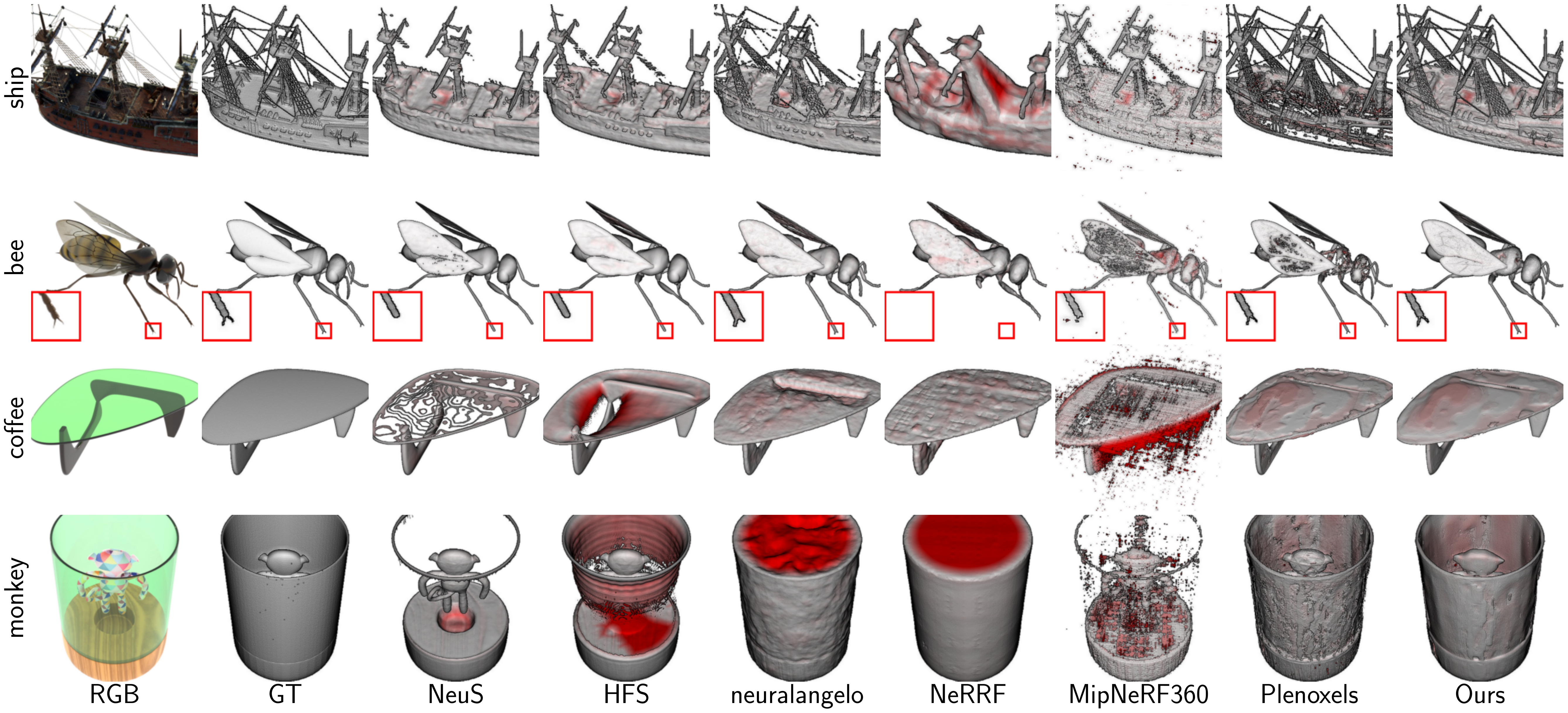}
\end{center}
   \caption{\textbf{Qualitative evaluation on synthetic datasets.} The red color indicates the L1 error in the reconstruction.
   Our method can reconstruct the translucent and thin surfaces missed in NeuS and HFS, while recovering more accurate and noise-free surfaces compared to neuralangelo and NeRF-based methods. Additional scenes can be found in the supplementary. 
   }
\label{fig:main_syn}
\end{figure*}


We quantitatively and qualitatively evaluate our method on an extended version of the NeRF synthetic dataset~\cite{nerf}, with 8 additional objects with delicate and thin structures, and 8 objects with translucent materials. We also show a qualitative comparison on challenging real-world scenes containing translucent surfaces. 
We compare with recent SDF optimization methods including NeuS~\cite{neus} and HFS~\cite{HFS},
and NeRF-based methods including Plenoxels~\cite{plenoxels}
and MipNeRF360~\cite{mipnerf360}.

\subsection{Datasets}

\noindent
\textbf{Synthetic Thin \& Translucent}: We render 8 scenes with thin structures and 8 with translucent surfaces with Blender~\cite{blender}. 
We sampled 100 different training views from a full sphere. For Thin dataset, we included the ``ficus" and ``ship" scenes from the original NeRF Synthetic dataset~\cite{nerf}. For each scene, we also rendered the depth and converted them to dense point clouds for quantitative geometry evaluation. This removes any invisible inner structures in the 3D assets. Except for the translucent scene ``monkey" and ``vase" where part of the object is completely surrounded by translucent surfaces, we hence directly extracted the scene meshes as the ground truth geometry. We will release all the datasets with reference geometry upon publication.

\noindent
\textbf{Real-World}: 
We additionally captured a real-world scene with thin structures and two with translucent surfaces to qualitatively evaluate our performance. The real-world captures are processed with Colmap~\cite{colmap1, colmap2} to obtain camera parameters. \ww{Note that since our work aims to resolve the geometry-material ambiguity in image-based neural reconstruction instead of handling complicated specular reflection or light transport, we hence focus on real-world thin translucent surfaces with less obvious view-dependent effects, such as cups.}

\begin{figure*}[t]
  \centering
  \begin{minipage}[]{0.6\textwidth}
  \includegraphics[width=\textwidth]{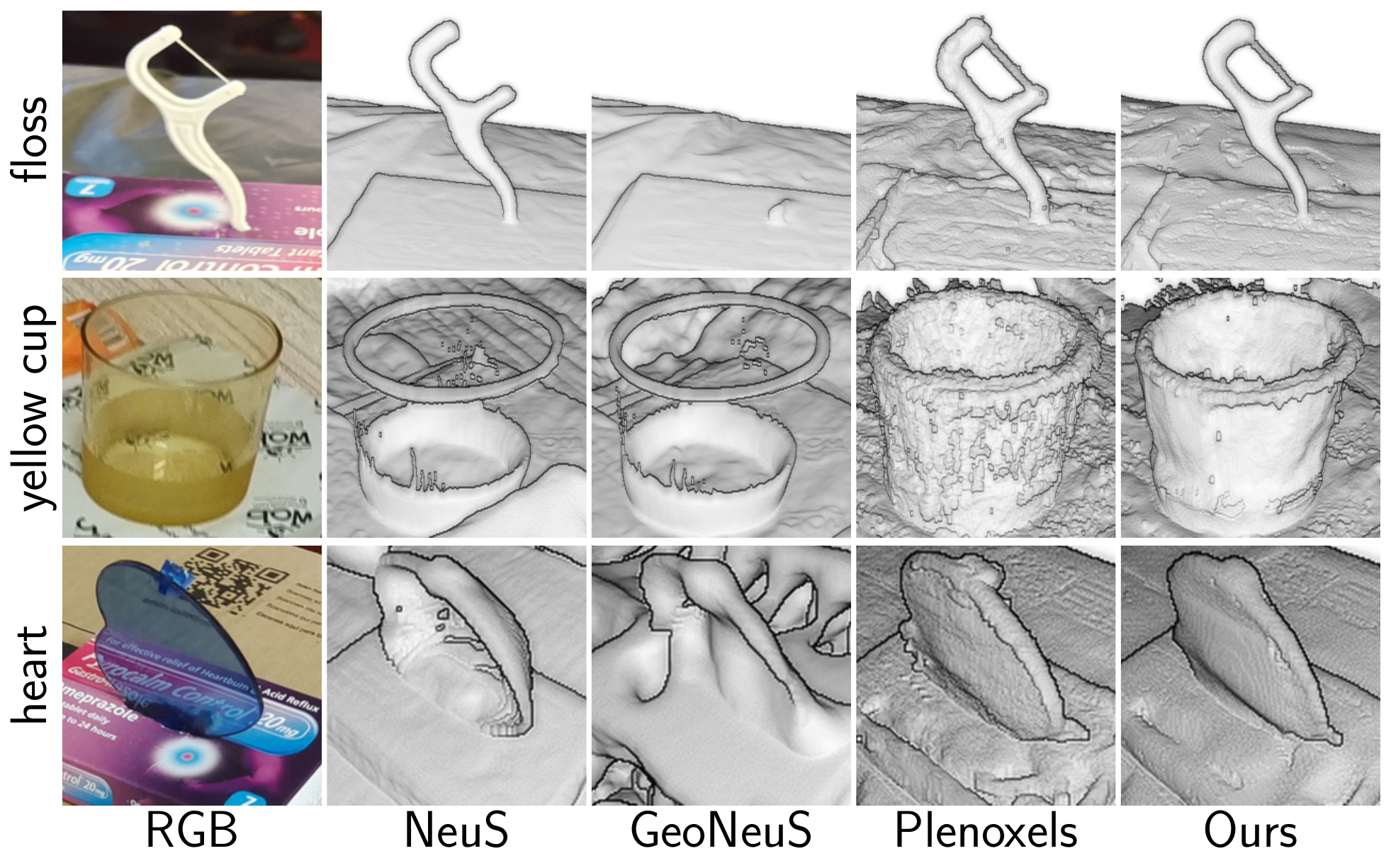}
   \caption{\textbf{Qualitative evaluation on translucent real-world scenes.} 
    The images are not aligned perfectly due to the use of NDC during optimization.
   }
\label{fig:llff}
  \end{minipage}
\hfill
  \begin{minipage}[]{0.35\textwidth}
\small
\centering
\singlespacing
\tabcolsep=0.06cm
%
%
\begin{tabular}{lrrrr}
\toprule
     name &  ship &  ficus & table &  monkey \\
\midrule
    no tv &    .317 &  .279 & .373 &   .777 \\
 notvnorm &    .373 &  .333 & .404 &   .970 \\
 no trunc &    .522 &  .837 & .431 &   1.03 \\
  no conv &    .287 &  .266 & .412 &   .857 \\
single lv &    .624 &  .583 & .470 &   .946 \\
     Ours &    \first{.277} &  \first{.240} & \first{.373} &   \first{.776} \\
\bottomrule
\end{tabular}
    \captionof{table}{\textbf{Quantitative results of ablation study.} We report the Chamfer distance $ \times 10^{-2}$ on two scenes from each synthetic dataset. More results can be found in the supplementary.
    }
    \label{tab:abla}
  \end{minipage}
\end{figure*}

\subsection{Baselines}

We compared our method with the state-of-the-art SDF-based reconstruction methods, as well as NeRF-based methods with level set geometry.
%
We compare with NeuS~\cite{neus} HFS~\cite{HFS}, GeoNueS~\cite{GeoNeus} and neuralangelo~\cite{neuralangelo}. 
Since GeoNueS~\cite{GeoNeus} requires SfM points and visibility masks as input, we only compare with it qualitatively in real-world scenes.
%
We also compare with Plenoxels~\cite{plenoxels} and MipNeRF~360~\cite{mipnerf360}, a follow-up of NeRF with conical frustum sampling and weight regularization. We also compare with NeRRF~\cite{chen2023nerrf}, which reconstructs fully transparent objects with additional mask supervision.

\subsection{Evaluation on Synthetic dataset}

We quantitatively evaluate our method on the synthetic datasets and report the Chamfer-L1 distance as evaluated in~\cite{dtu} in Table \ref{tab:chamfer}. 
HFS failed and learned empty surfaces on two Translucent scenes, while neuralangelo failed on one scene in Thin Blender.
In comparison, our method signifcantly outperforms the baselines, and the difference is particularly noticeable when compared to SDF-based methods on the translucent dataset, as the baselines cannot properly represent translucent surfaces. NeRF-based methods can potentially recover the translucent surfaces with a low density level set, but their ambiguity in representation leads to significant noise in the reconstruction.



We show qualitative comparison in Figure~\ref{fig:main_syn}. \ww{While NeuS, HFS and Neuralangelo are capable of reconstructing highly smooth surfaces, they cannot properly capture translucent or thin surfaces. 
Surfaces from Plenoxels and MipNeRF360 contain holes and floaters and are unsmooth.} NeRRF is capable of reconstructing smooth translucent surfaces, but fails to model intricate structures.
Our approach is capable of reconstructing surfaces with minimal artifacts. In Figure~\ref{fig:low_res}, we show surfaces reconstructed from 800, 400, and 200 resolution images on a scene with thin structures. Our method can reconstruct the thin surfaces despite the heavy blending effects in the low-resolution images.

\subsection{Evaluation on Real World Dataset}

We show the qualitative comparison on real-world scenes with thin and translucent surfaces in Figure~\ref{fig:llff}. 
\ww{
We show surfaces extracted Plenoxels using $\sigma=10$ level sets, as it has the best qualitative results compared to $30, 50$ levels.
Neus and GeoNueS fail to reconstruct the majority of the thin or translucent surface, while surfaces from Plenoxels are noisy and unsmooth.
\ww{Our method mitigates the artifacts inherited from NeRF initialization and reconstructs surfaces with much higher quality.} 
\ww{We would like to highlight that, due to the highly ill-posed nature of the problem and view-dependent appearance changes caused by global brightness shifts in an uncontrolled capture environment, it is nearly impossible to reconstruct the translucent surfaces perfectly. Our method achieves significant improvement upon baselines, validating the feasibility of our approach.}
\ww{Additional comparison with neuralangelo~\cite{neuralangelo} can be found in the supplementary.}
}

\subsection{Ablation}
\label{sec:abla}

As shown in Table~\ref{tab:abla},
the truncated alpha compositing deals with the inner volume artifacts inherited from initialization, while the surface regularization $\mathcal{L}_{\mathbf{n}_1}, \mathcal{L}_{\mathbf{n}_2}, \mathcal{L}_\delta$ and convergence loss $\mathcal{L}_c$ encourage smooth and accurate surfaces. Using a single level set $\bm{\tau}_\sigma=\{10\}$ to initialize the surface fails to capture all information in the pre-trained NeRF and causes artifacts in optimization. Ours with all techniques enabled achieves the best performance. More results can be found in the supplementary.

\section{Conclusion}

We present \name, a grid-based surface representation with decoupled geometry, opacity, and appearance. We develop closed-form intersection finding and differentiable alpha compositing to optimize the surface via photometric loss. 
Our representation utilizes initialization from efficient Plenoxels \cite{plenoxels}, and incorporates truncated rendering and additional surface regularizations to reconstruct high-quality surfaces for translucent objects and thin structures with heavy blending effects.



\noindent
\textbf{Limitation}: 
Compared to MLP-based SDF methods such as NeuS~\cite{neus} and neuralangelo~\cite{neuralangelo}, our reconstructed surface tends to be less smooth due to the lack of spatial smoothness encoded in the MLP; see Figure~\ref{fig:main_syn}. It presents a trade-off: stronger surface regularization can certainly give smoother surfaces, but can also destroy delicate thin structures in the reconstruction. 
\ww{In addition, we focus only on decoupling geometry and material ambiguity in existing volumetric representation, and do not specifically handle strong reflections as in Ref-NeRF~\cite{refnerf}. 
}

{
    \small
    \bibliographystyle{ieeenat_fullname}
    \bibliography{main}

\begin{thebibliography}{65}
\providecommand{\natexlab}[1]{#1}
\providecommand{\url}[1]{\texttt{#1}}
\expandafter\ifx\csname urlstyle\endcsname\relax
  \providecommand{\doi}[1]{doi: #1}\else
  \providecommand{\doi}{doi: \begingroup \urlstyle{rm}\Url}\fi

\bibitem[Adalsteinsson and Sethian(1995)]{sdf_redistance}
David Adalsteinsson and James~A. Sethian.
\newblock A fast level set method for propagating interfaces.
\newblock \emph{Journal of Computational Physics}, 118\penalty0 (2):\penalty0
  269--277, 1995.

\bibitem[Barron et~al.(2022)Barron, Mildenhall, Verbin, Srinivasan, and
  Hedman]{mipnerf360}
Jonathan~T. Barron, Ben Mildenhall, Dor Verbin, Pratul~P. Srinivasan, and Peter
  Hedman.
\newblock Mip-nerf 360: Unbounded anti-aliased neural radiance fields.
\newblock \emph{CVPR}, 2022.

\bibitem[Cai et~al.(2023)Cai, Huang, Jia, Lv, and Fu]{neuda}
Bowen Cai, Jinchi Huang, Rongfei Jia, Chengfei Lv, and Huan Fu.
\newblock Neuda: Neural deformable anchor for high-fidelity implicit surface
  reconstruction.
\newblock In \emph{Proceedings of the IEEE Conference on Computer Vision and
  Pattern Recognition}, 2023.

\bibitem[Chen et~al.(2023)Chen, Liu, Zhao, Zhou, and Zhang]{chen2023nerrf}
Xiaoxue Chen, Junchen Liu, Hao Zhao, Guyue Zhou, and Ya-Qin Zhang.
\newblock Nerrf: 3d reconstruction and view synthesis for transparent and
  specular objects with neural refractive-reflective fields, 2023.

\bibitem[Community(2018)]{blender}
Blender~Online Community.
\newblock \emph{Blender - a 3D modelling and rendering package}.
\newblock Blender Foundation, Stichting Blender Foundation, Amsterdam, 2018.

\bibitem[Compare(2020)]{CloudCompare}
Cloud Compare.
\newblock \emph{CloudCompare 3D point cloud and mesh processing software Open
  Source Project}, 2020.

\bibitem[Curless and Levoy(1996)]{tsdf}
Brian Curless and Marc Levoy.
\newblock A volumetric method for building complex models from range images.
\newblock In \emph{Proceedings of the 23rd Annual Conference on Computer
  Graphics and Interactive Techniques}, page 303–312, New York, NY, USA,
  1996. Association for Computing Machinery.

\bibitem[Deng et~al.(2022)Deng, Yang, Xiang, and Tong]{gram}
Yu Deng, Jiaolong Yang, Jianfeng Xiang, and Xin Tong.
\newblock Gram: Generative radiance manifolds for 3d-aware image generation.
\newblock In \emph{IEEE/CVF Conference on Computer Vision and Pattern
  Recognition}, 2022.

\bibitem[Detrixhe et~al.(2013)Detrixhe, Gibou, and Min]{fast_sweep2}
Miles Detrixhe, Frédéric Gibou, and Chohong Min.
\newblock A parallel fast sweeping method for the eikonal equation.
\newblock \emph{Journal of Computational Physics}, 237:\penalty0 46--55, 2013.

\bibitem[Edelsbrunner et~al.(1983)Edelsbrunner, Kirkpatrick, and
  Seidel]{alpha_shape}
H. Edelsbrunner, D. Kirkpatrick, and R. Seidel.
\newblock On the shape of a set of points in the plane.
\newblock \emph{IEEE Transactions on Information Theory}, 29\penalty0
  (4):\penalty0 551--559, 1983.

\bibitem[Fu et~al.(2022)Fu, Xu, Ong, and Tao]{GeoNeus}
Qiancheng Fu, Qingshan Xu, Yew-Soon Ong, and Wenbing Tao.
\newblock Geo-neus: Geometry-consistent neural implicit surfaces learning for
  multi-view reconstruction.
\newblock \emph{NeurIPS}, 2022.

\bibitem[Gu{\'e}don and Lepetit(2023)]{sugar}
Antoine Gu{\'e}don and Vincent Lepetit.
\newblock Sugar: Surface-aligned gaussian splatting for efficient 3d mesh
  reconstruction and high-quality mesh rendering.
\newblock \emph{arXiv preprint arXiv:2311.12775}, 2023.

\bibitem[{Hansson S\"oderlund} et~al.(2022){Hansson S\"oderlund}, Evans, and
  Akenine-M\"oller]{ray_trace_sdf}
Herman {Hansson S\"oderlund}, Alex Evans, and Tomas Akenine-M\"oller.
\newblock Ray tracing of signed distance function grids.
\newblock \emph{Journal of Computer Graphics Techniques (JCGT)}, 11\penalty0
  (3):\penalty0 94--113, 2022.

\bibitem[Haque et~al.(2023)Haque, Tancik, Efros, Holynski, and
  Kanazawa]{instructnerf2nerf}
Ayaan Haque, Matthew Tancik, Alexei Efros, Aleksander Holynski, and Angjoo
  Kanazawa.
\newblock Instruct-nerf2nerf: Editing 3d scenes with instructions.
\newblock In \emph{Proceedings of the IEEE/CVF International Conference on
  Computer Vision}, 2023.

\bibitem[He et~al.(2022)He, Sui, Huang, Dai, Lyu, and
  Liu]{laser_transparent_recon}
Kejing He, Congying Sui, Tianyu Huang, Rong Dai, Congyi Lyu, and Yun-Hui Liu.
\newblock 3d surface reconstruction of transparent objects using laser scanning
  with ltftf method.
\newblock \emph{Optics and Lasers in Engineering}, 148:\penalty0 106774, 2022.

\bibitem[Hinton()]{rmsprop}
Geoffrey Hinton.
\newblock Rmsprop.

\bibitem[Jensen et~al.(2014)Jensen, Dahl, Vogiatzis, Tola, and Aan{\ae}s]{dtu}
Rasmus Jensen, Anders Dahl, George Vogiatzis, Engil Tola, and Henrik Aan{\ae}s.
\newblock Large scale multi-view stereopsis evaluation.
\newblock In \emph{2014 IEEE Conference on Computer Vision and Pattern
  Recognition}, pages 406--413. IEEE, 2014.

\bibitem[Jiang et~al.(2023)Jiang, Jiang, Zhao, and Huang]{LEAP}
Hanwen Jiang, Zhenyu Jiang, Yue Zhao, and Qixing Huang.
\newblock Leap: Liberate sparse-view 3d modeling from camera poses.
\newblock \emph{ArXiv}, 2310.01410, 2023.

\bibitem[Jiang et~al.(2020)Jiang, Ji, Han, and Zwicker]{sdfdiff}
Yue Jiang, Dantong Ji, Zhizhong Han, and Matthias Zwicker.
\newblock Sdfdiff: Differentiable rendering of signed distance fields for 3d
  shape optimization.
\newblock In \emph{The IEEE/CVF Conference on Computer Vision and Pattern
  Recognition (CVPR)}, 2020.

\bibitem[Jzhangbs()]{python_dtu}
Jzhangbs.
\newblock Jzhangbs/dtueval-python: A fast python implementation of dtu mvs 2014
  evaluation.

\bibitem[Kazhdan et~al.(2006)Kazhdan, Bolitho, and Hoppe]{poisson_recon}
Michael Kazhdan, Matthew Bolitho, and Hugues Hoppe.
\newblock {Poisson Surface Reconstruction}.
\newblock In \emph{Symposium on Geometry Processing}. The Eurographics
  Association, 2006.

\bibitem[Kerbl et~al.(2023)Kerbl, Kopanas, Leimk{\"u}hler, and
  Drettakis]{gaussian_splatting}
Bernhard Kerbl, Georgios Kopanas, Thomas Leimk{\"u}hler, and George Drettakis.
\newblock 3d gaussian splatting for real-time radiance field rendering.
\newblock \emph{ACM Transactions on Graphics}, 42\penalty0 (4), 2023.

\bibitem[Kim et~al.(2022)Kim, Seo, and Han]{infonerf}
Mijeong Kim, Seonguk Seo, and Bohyung Han.
\newblock Infonerf: Ray entropy minimization for few-shot neural volume
  rendering.
\newblock In \emph{CVPR}, 2022.

\bibitem[Lee and Schachter(1980)]{delaunay}
D. Lee and B. Schachter.
\newblock Two algorithms for constructing a delaunay triangulation.
\newblock \emph{International Journal of Parallel Programming}, 9:\penalty0
  219--242, 1980.

\bibitem[Li et~al.(2022)Li, Yang, Zhai, Liu, Bao, and Zhang]{voxel_surf}
Hai Li, Xingrui Yang, Hongjia Zhai, Yuqian Liu, Hujun Bao, and Guofeng Zhang.
\newblock Vox-surf: Voxel-based implicit surface representation.
\newblock \emph{IEEE Transactions on Visualization and Computer Graphics},
  pages 1--12, 2022.

\bibitem[Li et~al.(2020)Li, Yeh, and Chandraker]{through_glass}
Zhengqin Li, Yu-Ying Yeh, and Manmohan Chandraker.
\newblock Through the looking glass: Neural 3d reconstruction of transparent
  shapes.
\newblock pages 1259--1268, 2020.

\bibitem[Li et~al.(2023{\natexlab{a}})Li, Long, Wang, Cao, Wang, Luo, and
  Xiao]{neto}
Zongcheng Li, Xiaoxiao Long, Yusen Wang, Tuo Cao, Wenping Wang, Fei Luo, and
  Chunxia Xiao.
\newblock Neto: Neural reconstruction of transparent objects with
  self-occlusion aware refraction-tracing.
\newblock \emph{arXiv:2303.11219}, 2023{\natexlab{a}}.

\bibitem[Li et~al.(2023{\natexlab{b}})Li, M\"uller, Evans, Taylor, Unberath,
  Liu, and Lin]{neuralangelo}
Zhaoshuo Li, Thomas M\"uller, Alex Evans, Russell~H Taylor, Mathias Unberath,
  Ming-Yu Liu, and Chen-Hsuan Lin.
\newblock Neuralangelo: High-fidelity neural surface reconstruction.
\newblock In \emph{CVPR}, 2023{\natexlab{b}}.

\bibitem[Liu et~al.(2023)Liu, Wu, Hoorick, Tokmakov, Zakharov, and
  Vondrick]{zero123}
Ruoshi Liu, Rundi Wu, Basile~Van Hoorick, Pavel Tokmakov, Sergey Zakharov, and
  Carl Vondrick.
\newblock Zero-1-to-3: Zero-shot one image to 3d object, 2023.

\bibitem[Liu et~al.(2020)Liu, Zhang, Peng, Shi, Pollefeys, and
  Cui]{liu2020dist}
Shaohui Liu, Yinda Zhang, Songyou Peng, Boxin Shi, Marc Pollefeys, and Zhaopeng
  Cui.
\newblock Dist: Rendering deep implicit signed distance function with
  differentiable sphere tracing.
\newblock In \emph{Proceedings of the IEEE/CVF Conference on Computer Vision
  and Pattern Recognition}, pages 2019--2028, 2020.

\bibitem[Long et~al.(2022)Long, Lin, Wang, Komura, and Wang]{sparse_neus}
Xiaoxiao Long, Cheng Lin, Peng Wang, Taku Komura, and Wenping Wang.
\newblock Sparseneus: Fast generalizable neural surface reconstruction from
  sparse views.
\newblock \emph{ECCV}, 2022.

\bibitem[Mildenhall et~al.(2020)Mildenhall, Srinivasan, Tancik, Barron,
  Ramamoorthi, and Ng]{nerf}
Ben Mildenhall, Pratul~P. Srinivasan, Matthew Tancik, Jonathan~T. Barron, Ravi
  Ramamoorthi, and Ren Ng.
\newblock Nerf: Representing scenes as neural radiance fields for view
  synthesis.
\newblock In \emph{ECCV}, 2020.

\bibitem[M\"uller et~al.(2022)M\"uller, Evans, Schied, and Keller]{instantNGP}
Thomas M\"uller, Alex Evans, Christoph Schied, and Alexander Keller.
\newblock Instant neural graphics primitives with a multiresolution hash
  encoding.
\newblock \emph{ACM Trans. Graph.}, 41\penalty0 (4):\penalty0 102:1--102:15,
  2022.

\bibitem[Niemeyer et~al.(2020)Niemeyer, Mescheder, Oechsle, and
  Geiger]{Niemeyer2020CVPR}
Michael Niemeyer, Lars Mescheder, Michael Oechsle, and Andreas Geiger.
\newblock Differentiable volumetric rendering: Learning implicit 3d
  representations without 3d supervision.
\newblock In \emph{Proceedings IEEE Conf. on Computer Vision and Pattern
  Recognition (CVPR)}, 2020.

\bibitem[Oechsle et~al.(2021)Oechsle, Peng, and Geiger]{UNISURF}
Michael Oechsle, Songyou Peng, and Andreas Geiger.
\newblock Unisurf: Unifying neural implicit surfaces and radiance fields for
  multi-view reconstruction.
\newblock In \emph{International Conference on Computer Vision (ICCV)}, 2021.

\bibitem[Park et~al.(2021)Park, Sinha, Barron, Bouaziz, Goldman, Seitz, and
  Martin-Brualla]{nerfies}
Keunhong Park, Utkarsh Sinha, Jonathan~T. Barron, Sofien Bouaziz, Dan~B
  Goldman, Steven~M. Seitz, and Ricardo Martin-Brualla.
\newblock Nerfies: Deformable neural radiance fields.
\newblock \emph{ICCV}, 2021.

\bibitem[Poole et~al.(2022)Poole, Jain, Barron, and Mildenhall]{dreamfusion}
Ben Poole, Ajay Jain, Jonathan~T. Barron, and Ben Mildenhall.
\newblock Dreamfusion: Text-to-3d using 2d diffusion.
\newblock \emph{arXiv}, 2022.

\bibitem[Press et~al.(2007)Press, Teukolsky, Vetterling, and
  Flannery]{numerical_recipes}
William~H. Press, Saul~A. Teukolsky, William~T. Vetterling, and Brian~P.
  Flannery.
\newblock \emph{Numerical Recipes 3rd Edition: The Art of Scientific
  Computing}.
\newblock Cambridge University Press, 3 edition, 2007.

\bibitem[Reiser et~al.(2021)Reiser, Peng, Liao, and Geiger]{kilonerf}
Christian Reiser, Songyou Peng, Yiyi Liao, and Andreas Geiger.
\newblock Kilonerf: Speeding up neural radiance fields with thousands of tiny
  mlps.
\newblock In \emph{International Conference on Computer Vision (ICCV)}, 2021.

\bibitem[Ren et~al.(2023)Ren, Wang, Zhang, Pollefeys, and Süsstrunk]{volrecon}
Yufan Ren, Fangjinhua Wang, Tong Zhang, Marc Pollefeys, and Sabine Süsstrunk.
\newblock Volrecon: Volume rendering of signed ray distance functions for
  generalizable multi-view reconstruction, 2023.

\bibitem[Rosu and Behnke(2022)]{hash_sdf}
Radu~Alexandru Rosu and Sven Behnke.
\newblock Hashsdf: Accurate implicit surfaces with fast local features on
  permutohedral lattices, 2022.

\bibitem[{Sara Fridovich-Keil and Alex Yu} et~al.(2022){Sara Fridovich-Keil and
  Alex Yu}, Tancik, Chen, Recht, and Kanazawa]{plenoxels}
{Sara Fridovich-Keil and Alex Yu}, Matthew Tancik, Qinhong Chen, Benjamin
  Recht, and Angjoo Kanazawa.
\newblock Plenoxels: Radiance fields without neural networks.
\newblock In \emph{CVPR}, 2022.

\bibitem[Sch\"{o}nberger and Frahm(2016{\natexlab{a}})]{colmap1}
Johannes~Lutz Sch\"{o}nberger and Jan-Michael Frahm.
\newblock Structure-from-motion revisited.
\newblock In \emph{Conference on Computer Vision and Pattern Recognition
  (CVPR)}, 2016{\natexlab{a}}.

\bibitem[Sch\"{o}nberger and Frahm(2016{\natexlab{b}})]{schoenberger2016sfm}
Johannes~Lutz Sch\"{o}nberger and Jan-Michael Frahm.
\newblock Structure-from-motion revisited.
\newblock In \emph{Conference on Computer Vision and Pattern Recognition
  (CVPR)}, 2016{\natexlab{b}}.

\bibitem[Sch\"{o}nberger et~al.(2016{\natexlab{a}})Sch\"{o}nberger, Zheng,
  Pollefeys, and Frahm]{colmap2}
Johannes~Lutz Sch\"{o}nberger, Enliang Zheng, Marc Pollefeys, and Jan-Michael
  Frahm.
\newblock Pixelwise view selection for unstructured multi-view stereo.
\newblock In \emph{European Conference on Computer Vision (ECCV)},
  2016{\natexlab{a}}.

\bibitem[Sch\"{o}nberger et~al.(2016{\natexlab{b}})Sch\"{o}nberger, Zheng,
  Pollefeys, and Frahm]{schoenberger2016mvs}
Johannes~Lutz Sch\"{o}nberger, Enliang Zheng, Marc Pollefeys, and Jan-Michael
  Frahm.
\newblock Pixelwise view selection for unstructured multi-view stereo.
\newblock In \emph{European Conference on Computer Vision (ECCV)},
  2016{\natexlab{b}}.

\bibitem[Schwarz et~al.(2020)Schwarz, Liao, Niemeyer, and Geiger]{graf}
Katja Schwarz, Yiyi Liao, Michael Niemeyer, and Andreas Geiger.
\newblock Graf: Generative radiance fields for 3d-aware image synthesis.
\newblock In \emph{Advances in Neural Information Processing Systems
  (NeurIPS)}, 2020.

\bibitem[Sethian(1996)]{sdf_redistance2}
James~A. Sethian.
\newblock A fast marching level set method for monotonically advancing fronts.
\newblock \emph{Proceedings of the National Academy of Sciences of the United
  States of America}, 93 4:\penalty0 1591--5, 1996.

\bibitem[Sethian(1999)]{sdf_redistance3}
J.~A. Sethian.
\newblock Fast marching methods.
\newblock \emph{SIAM Review}, 41\penalty0 (2):\penalty0 199--235, 1999.

\bibitem[Sullivan and Kaszynski(2019)]{pyvista}
C.~Bane Sullivan and Alexander Kaszynski.
\newblock {PyVista}: 3d plotting and mesh analysis through a streamlined
  interface for the visualization toolkit ({VTK}).
\newblock \emph{Journal of Open Source Software}, 4\penalty0 (37):\penalty0
  1450, 2019.

\bibitem[Sun et~al.(2022)Sun, Sun, and Chen]{directvoxgo}
Cheng Sun, Min Sun, and Hwann{-}Tzong Chen.
\newblock Direct voxel grid optimization: Super-fast convergence for radiance
  fields reconstruction.
\newblock In \emph{CVPR}, 2022.

\bibitem[Verbin et~al.(2022)Verbin, Hedman, Mildenhall, Zickler, Barron, and
  Srinivasan]{refnerf}
Dor Verbin, Peter Hedman, Ben Mildenhall, Todd Zickler, Jonathan~T. Barron, and
  Pratul~P. Srinivasan.
\newblock {Ref-NeRF}: Structured view-dependent appearance for neural radiance
  fields.
\newblock \emph{CVPR}, 2022.

\bibitem[Vicini et~al.(2022)Vicini, Speierer, and Jakob]{diffsdf}
Delio Vicini, Sébastien Speierer, and Wenzel Jakob.
\newblock Differentiable signed distance function rendering.
\newblock \emph{Transactions on Graphics (Proceedings of SIGGRAPH)},
  41\penalty0 (4):\penalty0 125:1--125:18, 2022.

\bibitem[Wang et~al.(2021)Wang, Liu, Liu, Theobalt, Komura, and Wang]{neus}
Peng Wang, Lingjie Liu, Yuan Liu, Christian Theobalt, Taku Komura, and Wenping
  Wang.
\newblock Neus: Learning neural implicit surfaces by volume rendering for
  multi-view reconstruction.
\newblock \emph{NeurIPS}, 2021.

\bibitem[Wang et~al.(2022{\natexlab{a}})Wang, Han, Habermann, Daniilidis,
  Theobalt, and Liu]{neus2}
Yiming Wang, Qin Han, Marc Habermann, Kostas Daniilidis, Christian Theobalt,
  and Lingjie Liu.
\newblock Neus2: Fast learning of neural implicit surfaces for multi-view
  reconstruction, 2022{\natexlab{a}}.

\bibitem[Wang et~al.(2022{\natexlab{b}})Wang, Skorokhodov, and Wonka]{HFS}
Yiqun Wang, Ivan Skorokhodov, and Peter Wonka.
\newblock Hf-neus: Improved surface reconstruction using high-frequency
  details.
\newblock \emph{arXiv preprint arXiv:2206.07850}, 2022{\natexlab{b}}.

\bibitem[Xin et~al.(2021)Xin, Qi, Ying, Hongdong, Xuan, and Qing]{hdrnerf}
Huang Xin, Zhang Qi, Feng Ying, Li Hongdong, Wang Xuan, and Wang Qing.
\newblock Hdr-nerf: High dynamic range neural radiance fields.
\newblock \emph{arXiv preprint arXiv:2111.14451}, 2021.

\bibitem[Xu et~al.(2022)Xu, Jiang, Wang, Fan, Shi, and Wang]{SinNeRF}
Dejia Xu, Yifan Jiang, Peihao Wang, Zhiwen Fan, Humphrey Shi, and Zhangyang
  Wang.
\newblock Sinnerf: Training neural radiance fields on complex scenes from a
  single image.
\newblock 2022.

\bibitem[Yariv et~al.(2020)Yariv, Kasten, Moran, Galun, Atzmon, Ronen, and
  Lipman]{idr}
Lior Yariv, Yoni Kasten, Dror Moran, Meirav Galun, Matan Atzmon, Basri Ronen,
  and Yaron Lipman.
\newblock Multiview neural surface reconstruction by disentangling geometry and
  appearance.
\newblock \emph{Advances in Neural Information Processing Systems}, 33, 2020.

\bibitem[Yariv et~al.(2021)Yariv, Gu, Kasten, and Lipman]{volsdf}
Lior Yariv, Jiatao Gu, Yoni Kasten, and Yaron Lipman.
\newblock Volume rendering of neural implicit surfaces.
\newblock In \emph{Thirty-Fifth Conference on Neural Information Processing
  Systems}, 2021.

\bibitem[Yu et~al.(2021{\natexlab{a}})Yu, Li, Tancik, Li, Ng, and
  Kanazawa]{plenoctrees}
Alex Yu, Ruilong Li, Matthew Tancik, Hao Li, Ren Ng, and Angjoo Kanazawa.
\newblock {PlenOctrees} for real-time rendering of neural radiance fields.
\newblock In \emph{ICCV}, 2021{\natexlab{a}}.

\bibitem[Yu et~al.(2021{\natexlab{b}})Yu, Ye, Tancik, and Kanazawa]{pixelnerf}
Alex Yu, Vickie Ye, Matthew Tancik, and Angjoo Kanazawa.
\newblock {pixelNeRF}: Neural radiance fields from one or few images.
\newblock In \emph{CVPR}, 2021{\natexlab{b}}.

\bibitem[Zhang et~al.(2020)Zhang, Riegler, Snavely, and Koltun]{nerf++}
Kai Zhang, Gernot Riegler, Noah Snavely, and Vladlen Koltun.
\newblock Nerf++: Analyzing and improving neural radiance fields.
\newblock \emph{arXiv:2010.07492}, 2020.

\bibitem[Zhang et~al.(2022)Zhang, Luan, Li, and Snavely]{iron}
Kai Zhang, Fujun Luan, Zhengqi Li, and Noah Snavely.
\newblock Iron: Inverse rendering by optimizing neural sdfs and materials from
  photometric images.
\newblock In \emph{IEEE Conf. Comput. Vis. Pattern Recog.}, 2022.

\bibitem[Zhao(2004)]{fast_sweep}
Hongkai Zhao.
\newblock A fast sweeping method for eikonal equations.
\newblock \emph{Math. Comput.}, 74:\penalty0 603--627, 2004.

\end{thebibliography}
}
\clearpage

\appendix
{
    \centering
    \Large
    \textbf{\name{}: Implicit Surface Reconstruction for Semi-Transparent and Thin Objects with Decoupled Geometry and Opacity} \\
    \vspace{0.5em}Supplementary Material \\
    \vspace{1.0em}
}

\section{Overview}

In the supplementary material, we include additional experiment details and evaluation results. 
We also encourage the reader to watch the video results contained in the supplementary files.



\section{Implementation Details}

\subsection{Closed-Form Intersection}

As briefly mentioned in Section~\ref{sec:rendering} of our paper, we determine the ray-surface intersections through the analytical solution of cubic polynomials. Note that a similar technique has been identified in previous works \cite{ray_trace_sdf, sdfdiff}, but they applied it on SDF only. We identify that the same approach can be applied to a more generalized implicit surface field without the Eikonal constraint. We now present the detailed derivation of it. 


Given a camera ray $\mathbf{r}(t) = \mathbf{o} + t \mathbf{d}$ with origin $\mathbf{o}$ and direction $\mathbf{d}$, our aim is to find the intersections between the ray and a level set surface with value $\tau_i$ within a voxel $v_\mathbf{x}$, which $\mathbf{r}(t)$ is guaranteed to hit. The value of the implicit surface field within $v_\mathbf{x}$ can be determined through the trilinear interpolation of eight surface scalars stored on the vertices of $v_\mathbf{x}$:
%
\begin{align}
\delta(\mathbf{x}) = &\trilearp(\mathbf{x}, \{\hat{\delta}_i\}_{i=1}^{8})
    \\
    = &(1-z)( (1-y)((1-x)\hat\delta_{1}+x\hat\delta_{5}) \nonumber 
    \\
    &+ y((1-x) \hat\delta_{3} + x \hat\delta_{7})  ) \nonumber
    \\
    &+ z( (1-y)((1-x) \hat\delta_{2}+x \hat\delta_{6})  \nonumber
    \\
    &+ y((1-x) \hat\delta_{4} + x \hat\delta_{8})  ) 
    \label{eq:trilearp_surf}
\end{align}

\noindent
where $[x,y,z] = \mathbf{x} - \mathbf{l}$ are the relative coordinates within the voxel, and $\mathbf{l} = \floor(\mathbf{x})$.
Note that $x,y,z \in [0,1]$. 
We first determine the near and far intersections $t_n, t_f$ between the ray and voxel $v_\mathbf{x}$ through the ray-box AABB algorithm, and then redefine a new camera origin $\mathbf{o}' = \mathbf{o} + t_n \mathbf{d} - \mathbf{l}$. 
We hence directly have $[x,y,z] = \mathbf{o}' + t' \mathbf{d} \in [0,1]$, where $t' = t - t_n$ without the need for calculating relative coordinates again. By denoting $\mathbf{o}' = [o'_x,o'_y,o'_z]$, $\mathbf{d} = [d_x, d_y, d_z]$, we substitute the above as well as $\delta(\mathbf{x}) = \tau_i$ into Equation~\ref{eq:trilearp_surf}:

\begin{align}
    \tau_i = &(1-(o'_z + t' d_z))( (1-(o'_y + t' d_y))
    \nonumber  \\
    &((1-(o'_x + t' d_x))\hat\delta_{1}+(o'_x + t' d_x)\hat\delta_{5}) \nonumber \\
        &+ (o'_y + t' d_y)((1-(o'_x + t' d_x)) \hat\delta_{3} + (o'_x + t' d_x) \hat\delta_{7})  ) 
        \nonumber \\
        &+ (o'_z + t' d_z)( (1-(o'_y + t' d_y))
        \nonumber \\
        &((1-(o'_x + t' d_x)) \hat\delta_{2}+(o'_x + t' d_x) \hat\delta_{6})  
        \nonumber \\
        &+ (o'_y + t' d_y)((1-(o'_x + t' d_x)) \hat\delta_{4} + (o'_x + t' d_x) \hat\delta_{8})  ) \,.
\end{align}

By re-arranging the equation, we obtain:

\begin{align}
    \tau_i = f_3 t'^3 + f_2 t'^2 + f_1 t' + f_0
\end{align}
 
 \noindent
 where 

\begin{equation}
\begin{split}
 f_0 = &(m_{00} (1-o'_y) + m_{01} (o'_y)) (1-o'_x) 
 \\ &+ (m_{10} (1-o'_y) + m_{11} (o'_y)) (o'_x)
\\   
 f_1 = & (m_{10} (1-o'_y) + m_{11} (o'_y)) d_x + k_{1} (o'_x)
 \\ &-(m_{00} (1-o'_y) + m_{01} (o'_y)) d_x + k_{0} (1-o'_x) 
 \\
 f_2 = &k_{1} d_x+h_{1} (o'_x)-k_{0} d_x+h_{0} (1-o'_x)
 \\
 f_3 = &h_{1} d_x-h_{0} d_x
\end{split}
\end{equation}

\noindent
and
\begin{equation}
\begin{split}
m_{00} =& \hat{\delta}_1   (1-o'_z) + \hat{\delta}_2   (o'_z)
\\
m_{01} =& \hat{\delta}_3   (1-o'_z) + \hat{\delta}_4   (o'_z)
\\
m_{10} =& \hat{\delta}_5   (1-o'_z) + \hat{\delta}_6   (o'_z)
\\
m_{11} =& \hat{\delta}_7   (1-o'_z) + \hat{\delta}_8   (o'_z)
\\
k_{0} =& (m_{01} d_y + d_z (\hat{\delta}_4 - \hat{\delta}_3) (o'_y)) 
\\ &- (m_{00} d_y - d_z (\hat{\delta}_2 - \hat{\delta}_1) (1-o'_y))
\\
k_{1} =& (m_{11} d_y + d_z (\hat{\delta}_8 - \hat{\delta}_7) (o'_y)) 
\\ &-(m_{10} d_y - d_z (\hat{\delta}_6 - \hat{\delta}_5) (1-o'_y))
\\
h_{0} =& d_y d_z (\hat{\delta}_4 - \hat{\delta}_3) -d_y d_z (\hat{\delta}_2 - \hat{\delta}_1)
\\
h_{1} =& d_y d_z (\hat{\delta}_8 - \hat{\delta}_7) -d_y d_z (\hat{\delta}_6 - \hat{\delta}_5) \,.
\end{split}
\end{equation}

Therefore, we obtain a cubic polynomial with a single unknown $t'$. Note that here we only sketch the main idea. For the actual implementation, we refer to~\cite{ ray_trace_sdf} which provides a more concise implementation that formulates the cubic polynomials with fewer operations through the use of fused-multiply-add. 

We then incorporate Vieta's approach~\cite{numerical_recipes} to solve the real roots for $t'$ in an analytic way. Namely, we first re-write the cubic polynomial as follows:

\begin{align}
\tau_i &= t'^3 + a t'^2 + b t' + c
\\
a &= \frac{f_2}{f_3}, b = \frac{f_1}{f_3}, c = \frac{f_0}{f_3} \,.
\end{align}

Then, compute:

\begin{align}
Q &= \frac{a^2 - 3b}{9}
\\
R &= \frac{2a^3 -9ab+27c}{54}  \,.
\end{align}

If $R^2 < Q^3$, we have three real roots given by:

\begin{align}
\theta &= \arccos(\frac{R}{\sqrt{Q^3}})
\\
t'_1 &= -2 \sqrt{Q} \cos (\frac{\theta}{3}) - \frac{a}{3}
\\
t'_2 &= -2 \sqrt{Q} \cos (\frac{\theta -2\pi}{3}) - \frac{a}{3}
\\
t'_3 &= -2 \sqrt{Q} \cos (\frac{\theta +2\pi}{3}) - \frac{a}{3}
\end{align}

where $t'_1 \leq t'_2 \leq t'_3$. This can be trivially seen from $0 \leq \theta \leq \pi$, $\sqrt{Q} \geq 0$ and $\cos (\frac{\theta}{3}) \geq \cos (\frac{\theta -2\pi}{3}) \geq \cos (\frac{\theta +2\pi}{3})$.

If $R^2 \geq Q^3$, we only have a single real root. First compute:

\begin{align}
A &= -\sign(R) \left( |R| + \sqrt{R^2 - Q^3} \right)^{1/3}
\\
B &=
\begin{cases}
  Q/A, & \text{if}\ A \neq 0 \\
  0, & \text{otherwise}
\end{cases} \,.
\end{align}

Then, the only real root can be obtained as:

\begin{align}
    t'_1 = (A+b) - \frac{a}{3} \,.
\end{align}

The intersection coordinate can therefore be determined as $\mathbf{r}(t_n + t')$. We then check the intersections against the bounding box of each voxel $v_\mathbf{x}$ to remove any samples outside of the voxels.
Besides, the cubic polynomial might return multiple valid real roots within the voxel if $R^2 < Q^3$. If the roots are unique, that means the ray intersects with the same surface multiple times within the voxel, all the intersections are taken for rendering. However, if the roots are identical, we remove the redundant ones to prevent using the same intersection multiple times. 

As both the formulation of cubic polynomials and Vieta's approach are fully differentiable, we hence directly have gradients defined on our surface representation $\hat{\delta}$ from the photometric loss. 



\subsection{Hyperparameters}

\begin{figure}[t]
\begin{center}
  \includegraphics[width=0.45\textwidth]{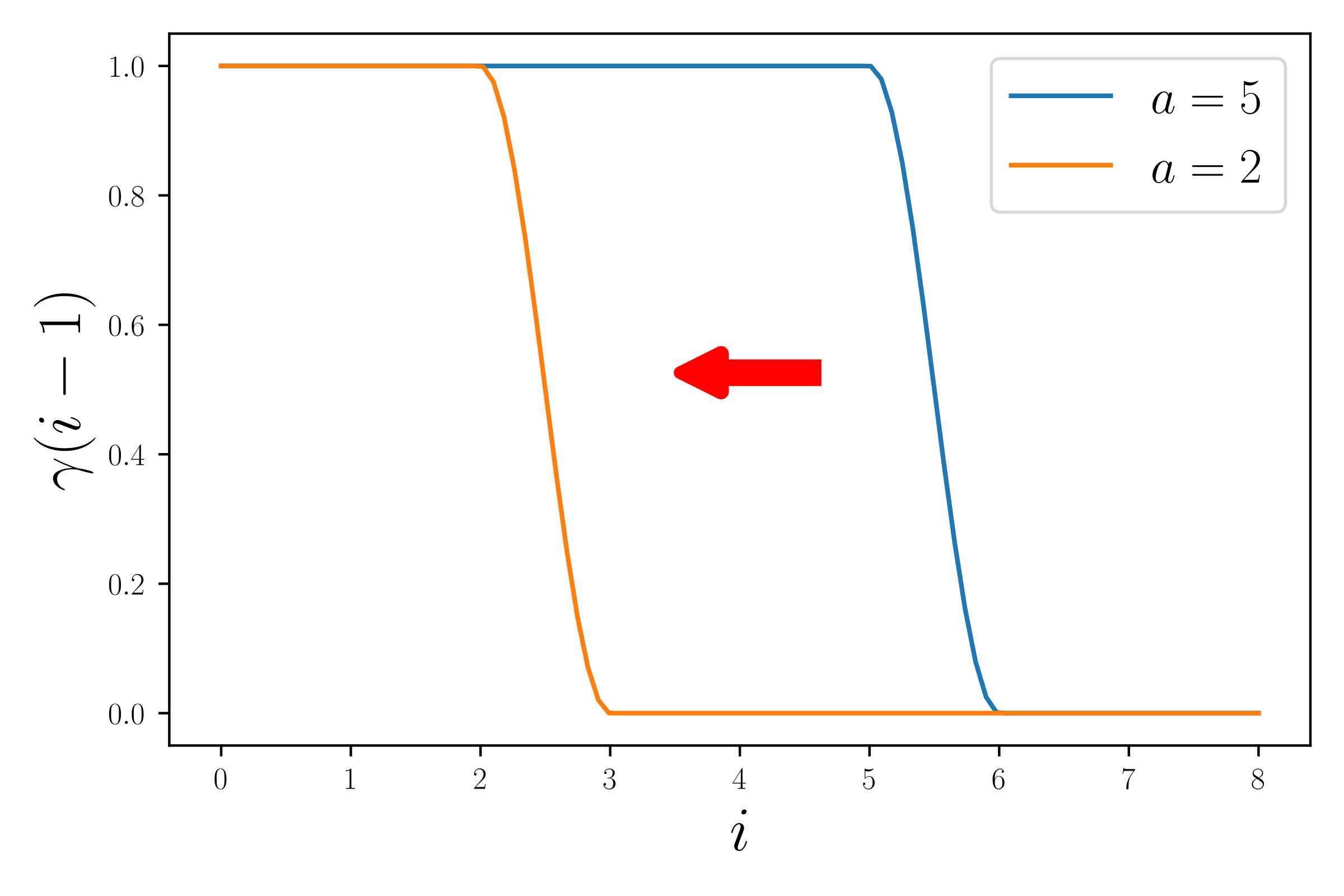}
\end{center}
    \caption{\textbf{The truncated alpha compositing reweight function $\gamma(i-1)$.} The x-axis is the index of the intersection starting from 1 (excluding intersections on back-facing surfaces). By reducing $a$, we effectively slide the curve to the left.}
    \label{fig:truncate_func}
\end{figure}

\begin{figure*}[t]
\begin{center}
  \includegraphics[width=0.95\textwidth]{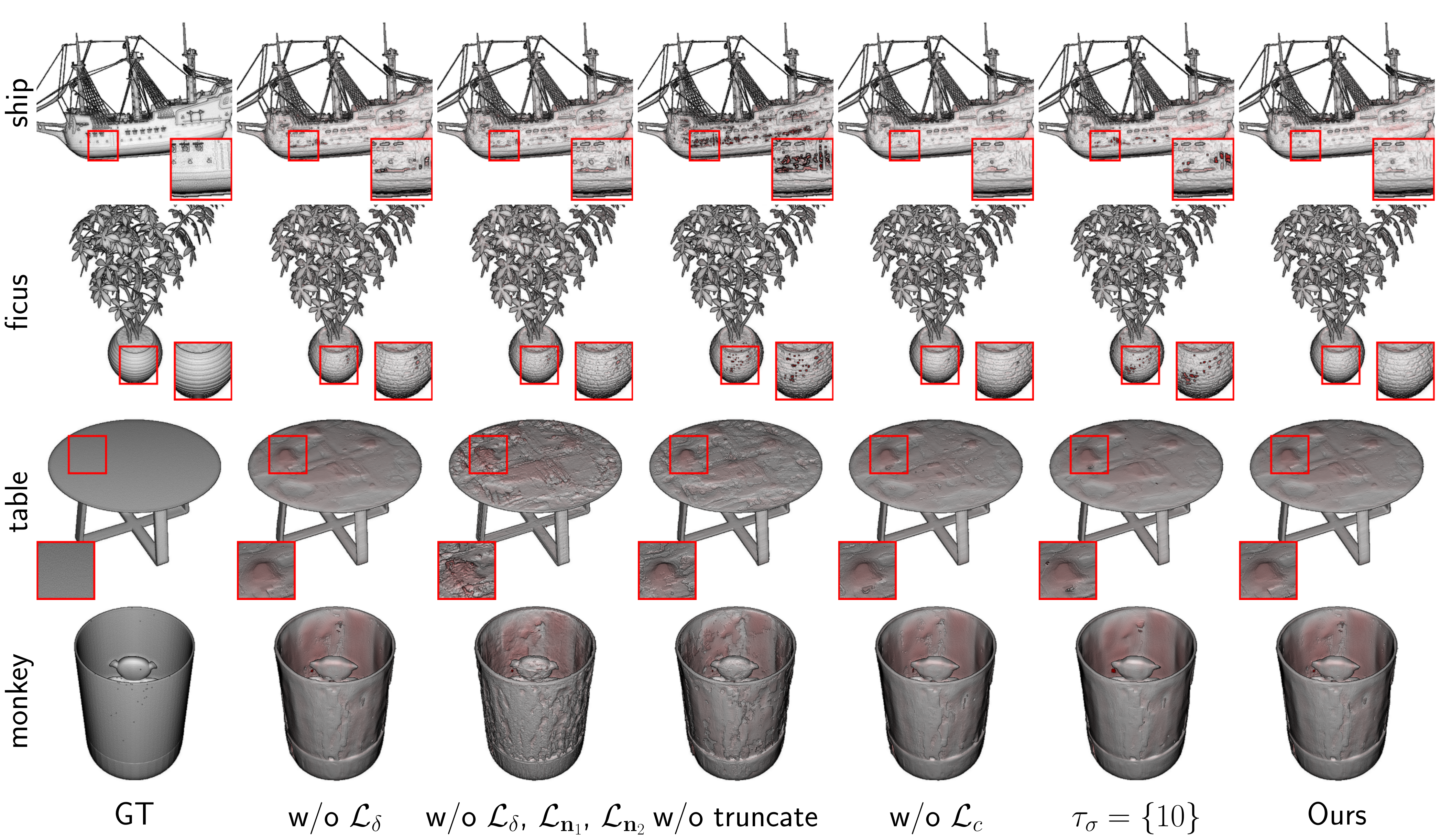}
\end{center}
   \caption{\textbf{Ablation Study.} We show the qualitative results of different ablations. Our full approach achieves the best quality overall.}
\label{fig:abla}
\end{figure*}

\begin{figure}[t]
\begin{center}
  \includegraphics[width=0.45\textwidth]{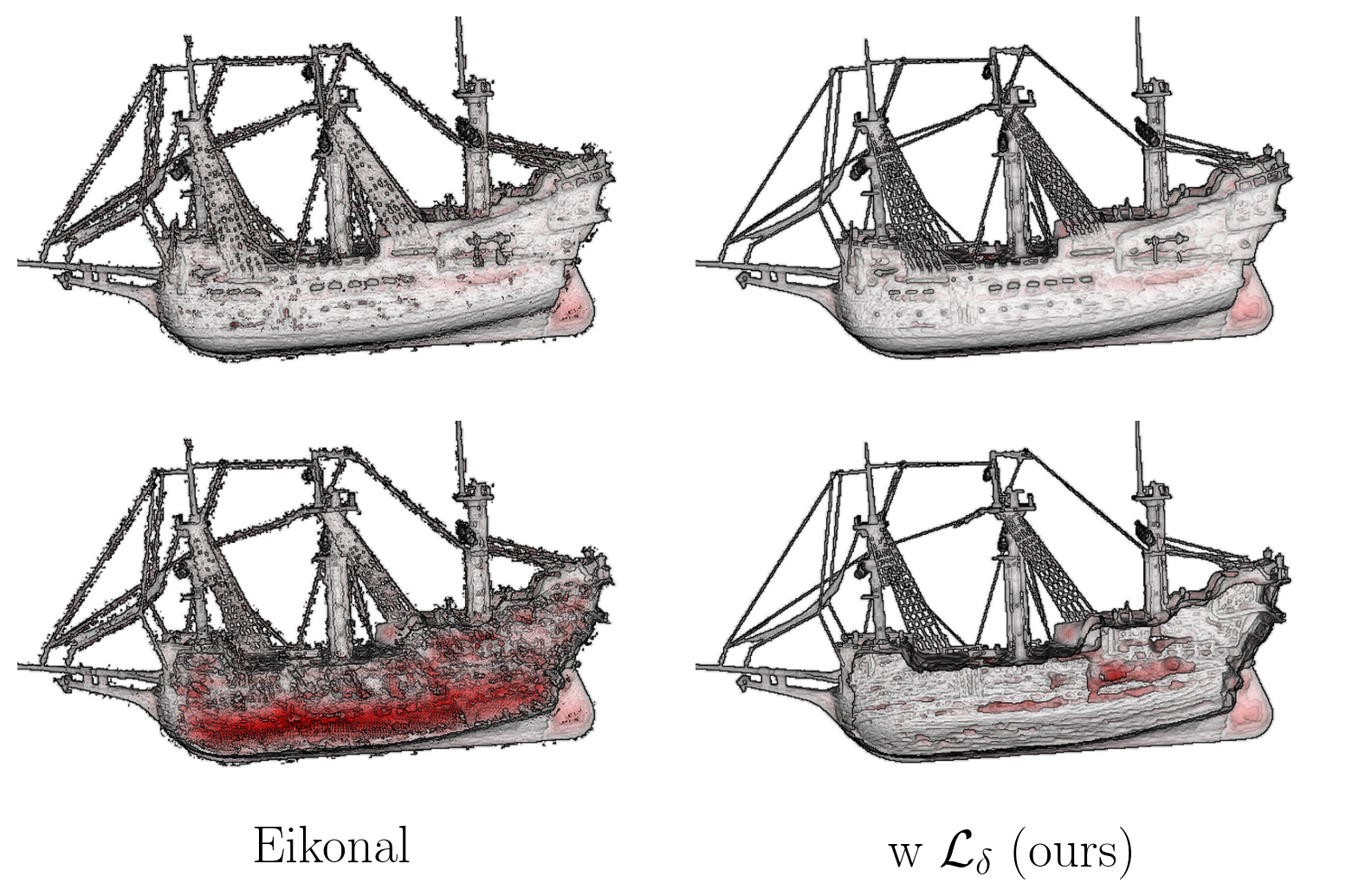}
\end{center}
   \caption{\textbf{Comparison with the Eikonal constraint regularization.} We visualize the out view (first column) and the inside view (second column) by cropping the surfaces along the y-axis. It can be clearly seen that the Eikonal constraint does not regularize the surface to be clean and smooth, but rather creates additional noises in optimization.}
\label{fig:abla_ek}
\end{figure}

\begin{figure}
    \centering
    \includegraphics[width=0.45\textwidth]{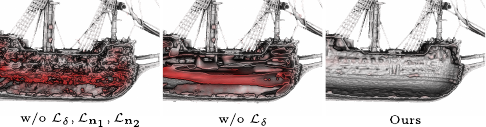}
    \caption{Reconstruction without regularization tends to give an enormous amount of redundant inner surfaces. Encouraging consistent normals in the local neighborhood via $\mathcal{L}_{\mathbf{n}_1}, \mathcal{L}_{\mathbf{n}_2}$ enforces smoother surfaces, but redundant surfaces still remain. With both normal regularization and TV loss applied on the surface scalar field, we can obtain clean and smooth reconstruction.}
    \label{fig:abla_crop}
\vspace{-15pt}
\end{figure}

Our code is based on Plenoxels~\cite{plenoxels}. 
We similarly use a sparse voxel grid of size $512^3$ where each vertex stores the surface scalar $\delta$, raw opacity $\sigma_\alpha$ and 9 SH coefficients for each color channel. We directly initialize all the grid values from Plenoxels pre-trained with original hyperparameters and prune voxels with densities $\sigma$ lower than $5$.
We use $s_\sigma=0.05$ to downscale the density values during initialization.
We train for $50k$ iterations with a batch size of $5k$ rays, which takes around 17 minutes for synthetic scenes and 22 minutes for real-world scenes on an NVIDIA A100-SXM-80GB GPU (excluding Plenoxels training). We used grid search to determine the optimal hyperparameters. We use the same delayed exponential learning rate schedule, where the learning rate is delayed with a scale of $0.01$ during first $25k$ iterations. 
As previously mentioned, the interval of level values is selected by first determining a valid range of Plenoxels density field. We then select a suitable number of level values, i.e., the carnality $n$ of our multi level sets, by trying $1, 3, 5, 10$ evenly-spaced level values on the ``ship" scene from NeRF Synthetic dataset. We found $n=5$ to give the best performance.
At the end of the training, we also remove invisible surfaces with opacity $\alpha$ less than $0.1$.

\paragraph{Synthetic}
For experiments on synthetic datasets, we initialize $5$ level sets at $\bm{\tau}_\sigma = \{10, 30, 50, 70, 90\}$, and linearly decay the truncated alpha compositing parameter $a$ from $5$ to $2$ in first $10k$ iterations. 
For surface scalars $\hat{\delta}$, we use $10^{-5}$ as both starting and end learning rate. For raw opacity values $\sigma_\alpha$, we start with $10^{-2}$ and end with $10^{-3}$. For SH we keep the learning rate at $10^{-3}$ without exponential decay or initial delay. We use the RMSProp~\cite{rmsprop} optimizer for training. 
For the regularization weights, we set $\lambda_c = 10^{-6}$ for the first $10k$ iterations and $0$ for the rest of training. \ww{We use 
$\lambda_\delta=10^{-3}$, 
$\lambda_\mathcal{H} = 10^{-4}$, 
$\lambda_\alpha = 10^{-9}$,
$\lambda_{\mathbf{n}_1} = 10^{-6}$ and $\lambda_{\mathbf{n}_2} = 0$ for the Thin dataset. $\lambda_{\mathbf{n}_2}$ was disabled as it tends to destroy the thin structures with rapid normal variations. For the Translucent dataset, we use $\lambda_\delta=10^{-5}$, $\lambda_\alpha = 10^{-11}$,  
$\lambda_\mathcal{H} = 10^{-4}$,
$\lambda_{\mathbf{n}_2} = 10^{-4}$, and linearly decay $\lambda_{\mathbf{n}_1}$ from $10^{-2}$ to $10^{-4}$.}

\paragraph{Real-World}
\ww{For experiments on real-world scenes, we initialize with less level sets $\bm{\tau}_\sigma = \{10, 30, 50\}$, as we found level values above 50 give almost empty surfaces due to higher density regularization in Plenoxels initialization. 
We use the hyperparameters used by the original authors to run LLFF experiments to train Plenoxels. For our method, we use level sets $\tau_\sigma = \{10, 30, 50\}$ as level value above $50$ gives almost empty space. We change the surface scalar learning rate to start and end both at $10^{-4}$ with a delay ratio of $10^{-2}$ and delay steps of $25k$. The learning rates of opacity and SH are the same as in the synthetic experiments.
For regularizations, we use the same $\lambda_c$ and $\lambda_\mathcal{H}$ as synthetic experiments, and set 
$\lambda_\delta=5 \times 10^{-3}$, $\lambda_\alpha = 10^{-9}$,  
$\lambda_{\mathbf{n}_2} = 10^{-3}$ and linearly decay $\lambda_{\mathbf{n}_1}$ from $10^{-2}$ to $10^{-3}$.
%
}


 For the implementation of surface TV loss $\mathcal{L}_\delta$, we calculate the gradient via forward finite difference in the same way as Plenoxels~\cite{plenoxels}:

\begin{align}
\nabla_x \hat{\delta}(i,j,k) = \frac{| \hat{\delta}(i+1,j,k) - \hat{\delta}(i,j,k)|D_x}{256}
\end{align}

\noindent 
where $i,j,k$ are the vertex coordinate, $D_x$ is the grid resolution in $x$ dimension and is $512$ for all experiments in our case.  $\nabla_y \hat{\delta}(i,j,k)$ and $\nabla_z \hat{\delta}(i,j,k)$ are calculated accordingly. We simply ignore the edge vertices when computing the surface TV loss by using the Neumann boundary conditions.

The truncated alpha compositing reweight function can be seen as a truncated Hann window \cite{nerfies}, as shown in Figure~\ref{fig:truncate_func}. By reducing $a$ during the training, we slide the curve to the left and hence gradually anneal the influence of later intersections.


\section{Additional Experiments}

\subsection{Synthetic Dataset}

\paragraph{Experiment Details}

For quantitative evaluation, we adapt the Python version of DTU~\cite{dtu} evaluation script~\cite{python_dtu}, where we extracted dense point clouds from all level surfaces and downsampled both predicted and ground truth points with 0.001 density before computing the Chamfer distance.
For evaluation of NeuS~\cite{neus} and HFS~\cite{HFS}, we first extracted the mesh using marching cubes with resolution $512^3$, then used the script to sample points on the mesh to compute the Chamfer distance. For evaluation of Plenoxels~\cite{plenoxels}, MipNeRF360~\cite{mipnerf360} and our method, we directly sample points on the implicit surfaces by sending dense virtual rays within each grid of a $512^3$ voxel grid through our closed-form intersection finding. This makes the computation of sample opacity and trimming of the surface easier. 

For training of NeuS \cite{neus}, HFS \cite{HFS} and MipNeRF360 \cite{mipnerf360}, we used the provided hyperparameters. We used the hyperparameters for real-world thin structure reconstruction experiments for NeuS, as we found it gives better performance on the NeRF Synthetic dataset. For training on the Translucent Blender dataset, we set the background to white for all methods as the semi-transparent objects are rendered with a white background in Blender.

\begin{figure*}[t]
\begin{center}
  \includegraphics[width=0.95\textwidth]{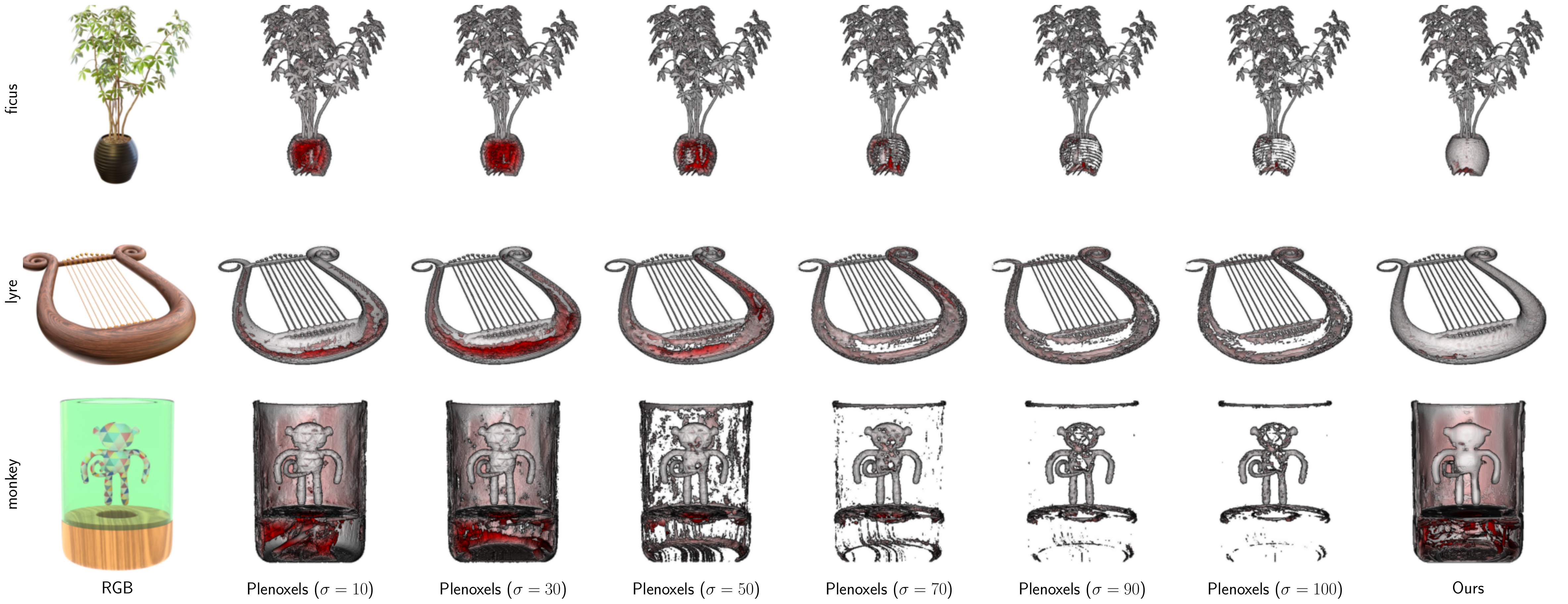}
  \includegraphics[width=0.95\textwidth]{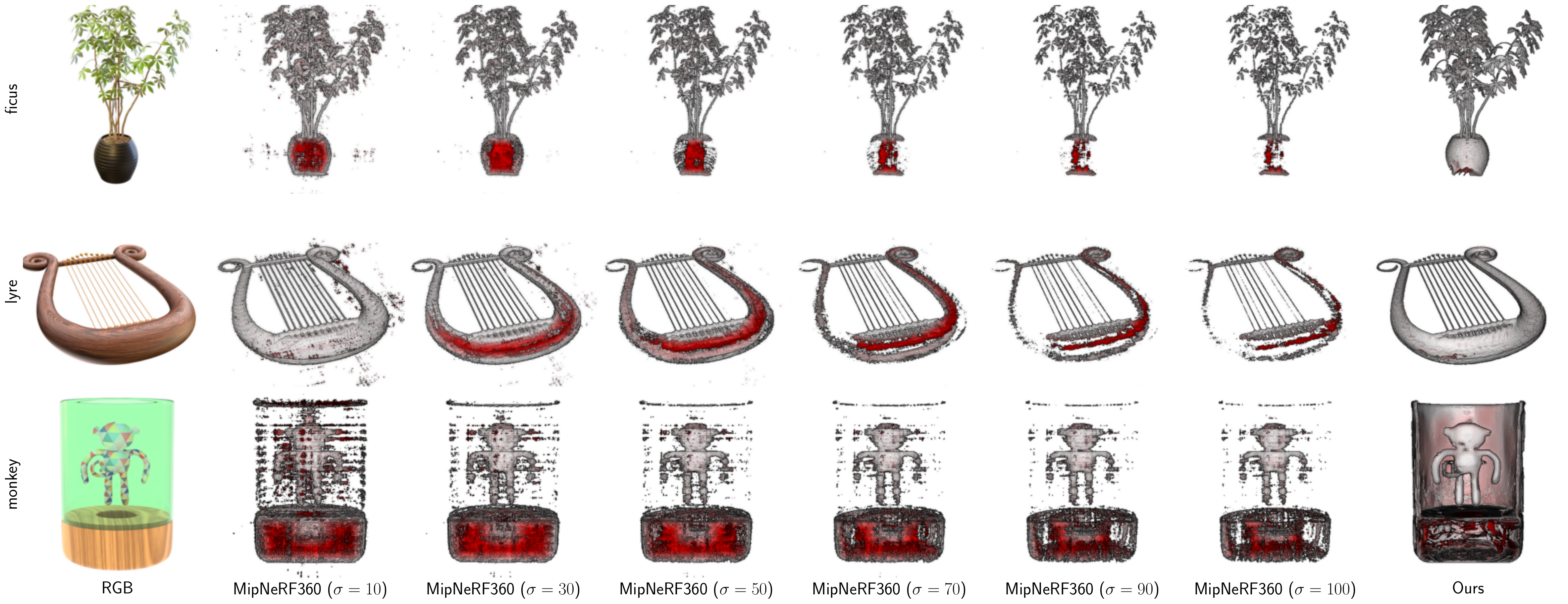}
\end{center}
    \caption{\textbf{Surfaces extracted using different level sets from Plenoxels~\cite{plenoxels} and MipNeRF360~\cite{mipnerf360}.} We remove part of the exterior surface in each scene to visualize the interior reconstructions. Due to the ambiguity of density representation, a low density level set gives more complete surfaces but could contain a significant amount of noise, whereas a high density level set can miss a lot of surfaces.}
    \label{fig:nerf_lvs}
\end{figure*}

\ww{To select a level set value on the density field of Pleboxels~\cite{plenoxels} and MipNeRF 360~\cite{mipnerf360} for surface extraction, we use the same methods as in~\cite{neus, UNISURF}, where we extracted and evaluated surfaces on levels $\tau_\sigma = \{10, 30, 50, 70, 90, 100\}$, which fully covers the surfaces we used to initialize from Plenoxels. We computed the average norm on Synthetic, Thin, and Translucent datasets and selected the level set value with the best Chamfer distance on each of the datasets. For Plenoxels, the level sets are $90, 30$ and for MipNeRF 360, the level sets are $50, 50$ for the two datasets respectively. We report the quantitative results for each level set in Tab~\ref{tab:nerf_lvs} and show a few qualitative examples in Figure~\ref{fig:nerf_lvs}.}

\begin{table}[]
\small
\centering
\singlespacing
\tabcolsep=0.06cm
\begin{tabular}{lccc}
\toprule
{} &   Thin & Translucent &  average \\
\midrule
Plen ($\sigma=10$)    &  0.759 &       0.813 &    0.786 \\
Plen ($\sigma=30$)    &  0.886 &       \first{0.761} &    0.824 \\
Plen ($\sigma=50$)    &  0.687 &       0.812 &    \first{0.750} \\
Plen ($\sigma=70$)    &  0.563 &       1.062 &    0.812 \\
Plen ($\sigma=90$)    &  \first{0.526} &       1.597 &    1.062 \\
Plen ($\sigma=100$)   &  0.541 &       1.832 &    1.186 \\
Mip360 ($\sigma=10$)  &  1.882 &        3.76 &    2.821 \\
Mip360 ($\sigma=30$)  &  1.468 &       3.081 &    2.274 \\
Mip360 ($\sigma=50$)  &  \first{1.445} &       \first{3.063} &    \first{2.254} \\
Mip360 ($\sigma=70$)  &  1.526 &        3.07 &    2.298 \\
Mip360 ($\sigma=90$)  &  1.635 &       3.116 &    2.376 \\
Mip360 ($\sigma=100$) &  1.693 &       3.203 &    2.448 \\
\bottomrule
\end{tabular}
    \caption{\textbf{Chamfer distance $\downarrow \times 10^{-2}$ on synthetic datasets.}
    We color the \colorbox{first}{best} level sets for Plenoxels~\cite{plenoxels} (Plen in table) and MipNeRF360~\cite{mipnerf360} (Mip360 in table) respectively.
    }
    \label{tab:nerf_lvs}
\end{table}

\paragraph{Additional Results}

\begin{figure*}[p]
\begin{center}
  \includegraphics[width=0.95\textwidth]{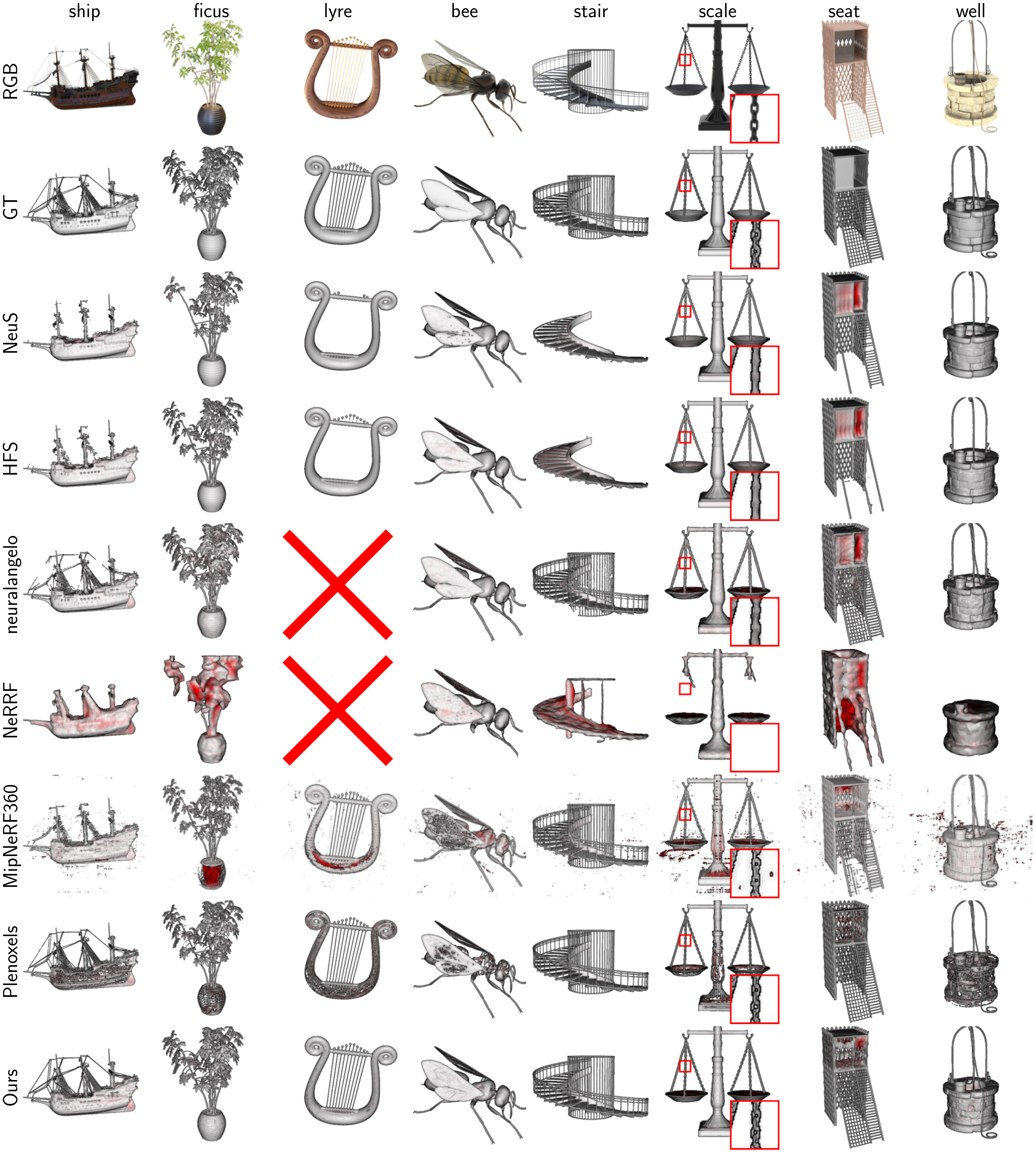}
\end{center}
   \caption{\textbf{Qualitative results on Thin Blender dataset.} 
   Note that neuralangelo \cite{neuralangelo} failed to learn any surface on the ``lyre" scene.
   }
\label{fig:delicate_all}

\centering
\small
\centering
\singlespacing
\begin{tabular}{lrrrrrrrrr}
\toprule
             {} &  ship &  ficus &  lyre &   bee &  stair &  scale &  seat &  well &   avg \\
\midrule
  Plen ($\sigma=90$) &    0.476 &  0.431 & \second{0.522} & 0.541 &  \second{0.206} &  0.884 & 0.497 & 0.647 & 0.526 \\
Mip360 ($\sigma=50$) &    1.217 &  2.640 & 1.311 & 1.181 &  0.279 &  2.243 & 0.744 & 1.941 & 1.445 \\
NeuS &    0.552 &  1.667 & 0.812 & \second{0.242} &  4.087 &  \first{0.237} & 0.431 & 0.360 & 1.049 \\
HFS &    0.514 &  \second{0.374} & 0.781 & 0.336 &  4.316 &  \second{0.271} & \second{0.425} & 0.385 & 0.925 \\
neuralangelo &    \first{0.270} &  0.411 &   NaN & 0.377 &  0.426 &  0.789 & 0.432 & \first{0.266} & \second{0.424} \\
NeRRF &            1.74 &           2.899 &             NaN &           1.164 &           3.731 &           1.359 &            2.17 &           3.381 &           2.349 \\
Ours &    \second{0.277} &  \first{0.240} & \first{0.188} & \first{0.207} &  \first{0.176} &  0.575 & \first{0.288} & \second{0.319} & \first{0.284} \\
\bottomrule
\end{tabular}

    \captionof{table}{\textbf{Chamfer distance $\downarrow \times 10^{-2}$ on Thin Blender datasets.} We color the \colorbox{first}{best}, \colorbox{second}{second best} methods. 
    }
    \label{tab:delicate}
\end{figure*}

\begin{figure*}[p]
\begin{center}
  \includegraphics[width=0.95\textwidth]{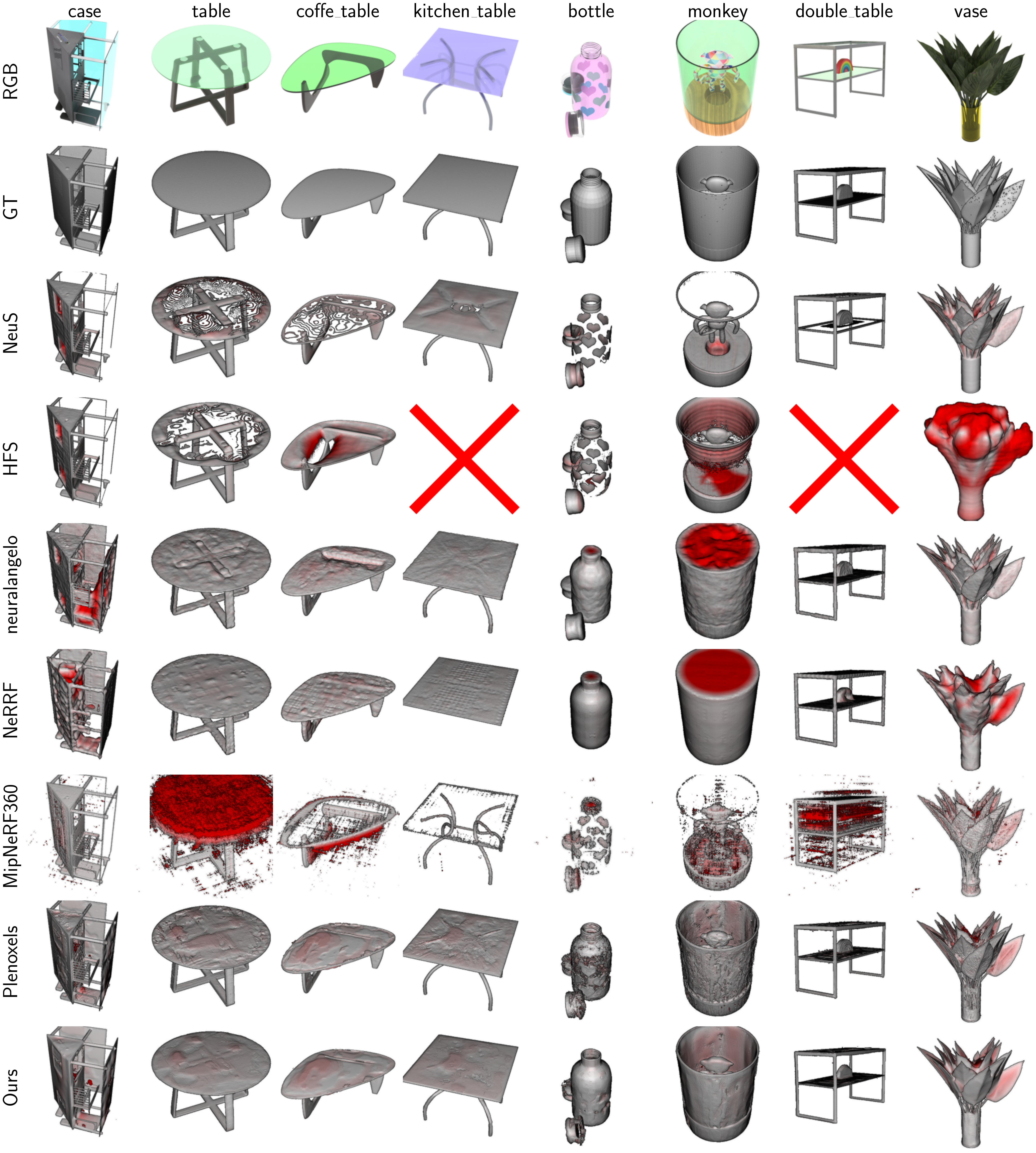}
\end{center}
   \caption{\textbf{Qualitative results on Translucent Blender dataset.} Note that HFS \cite{HFS} fails to learn any surface on ``kitchen table" and ``double table" scenes. We removed some exterior surfaces in the ``monkey" scene to show the interior surfaces.}
\label{fig:transparent_all}


\small
\centering
\singlespacing
\begin{tabular}{lrrrrrrrrr}
\toprule
             name &  case &  table &  coffee &  kitchen &  bottle &  monkey &  double &  vase &   avg \\
\midrule
  Plen ($\sigma=30$) & \second{1.195} &          \second{0.438} &        \first{0.706} &          \second{0.378} &     0.709 &   \second{1.084} &         0.557 &      \second{1.024} & \second{0.761} \\
Mip360 ($\sigma=50$) & 4.012 &          5.217 &        2.096 &          2.375 &     2.324 &   4.390 &         2.900 &      1.190 & 3.063 \\
NeuS & 5.091 &          1.188 &        1.070 &          0.392 &     2.271 &   5.923 &         0.874 &      1.946 & 2.344 \\
HFS & 5.094 &          0.854 &        2.839 &            NaN &     1.493 &   3.223 &           NaN &      8.684 & 3.698 \\
neuralangelo & 2.072 &          0.483 &        0.798 &          0.577 &     \first{0.395} &   2.464 &         \second{0.513} &      1.701 & 1.125 \\
NeRRF &           1.267 &           0.571 &           0.822 &           3.091 &            3.72 &           2.905 &           0.829 &           3.486 &           2.086 \\
             Ours & \first{0.835} &          \first{0.373} &        \second{0.717} &          \first{0.255} &     \second{0.653} &   \first{0.776} &         \first{0.512} &      \first{0.870} & \first{0.624} \\
\bottomrule
\end{tabular}
    \captionof{table}{\textbf{Chamfer distance $\downarrow \times 10^{-2}$ on Semi-Transparent Blender datasets.} We color the \colorbox{first}{best}, \colorbox{second}{second best} methods. Note that HFS \cite{HFS} fails to learn any surface on ``kitchen table" and ``double table" scenes.}
    \label{tab:semi_t}
\end{figure*}

\begin{figure*}[p]
\begin{center}
  \includegraphics[width=0.95\textwidth]{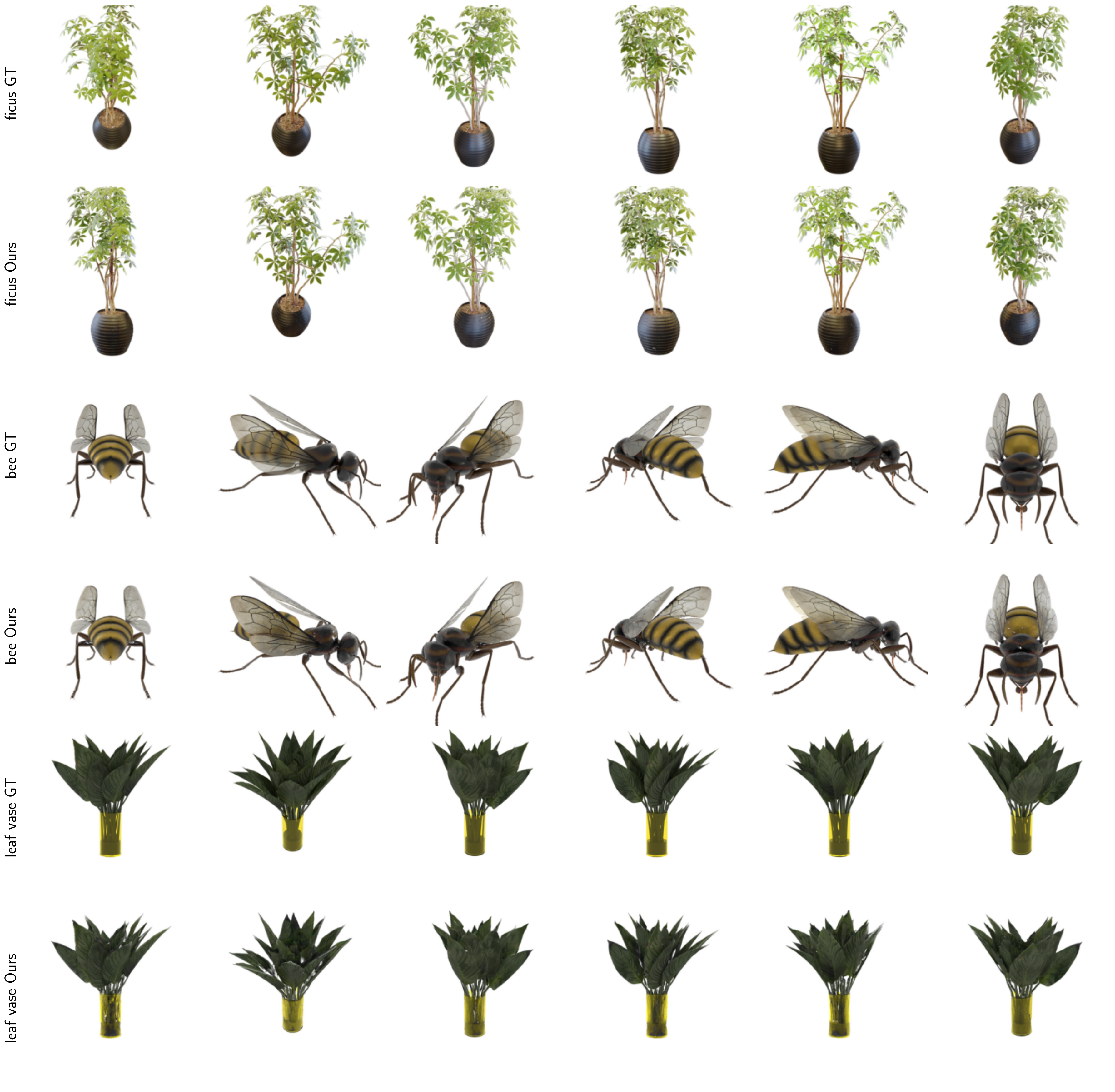}
\end{center}
   \caption{\textbf{RGB renderings of our methods on synthetic datasets.}}
\label{fig:rgb_renderings}
\end{figure*}

We show all the qualitative results in 
\ref{fig:delicate_all}, \ref{fig:transparent_all}, as well as the individual Chamfer distance for each scene in 
\ref{tab:delicate} and \ref{tab:semi_t}.
The qualitative comparisons shown in both main paper and the supplementary are done by first evaluating the L1 error on each sampled point, then rendering the point cloud with Eye-Dome Lighting (EDL) using PyVista~\cite{pyvista}.
We also show additional novel view RGB renderings of our method in Figure~\ref{fig:rgb_renderings}. But please note that we do not claim state-of-the-art performance in novel view synthesis.

\subsection{Real-World Dataset}

\begin{figure}[t]
\begin{center}
  \includegraphics[width=0.45\textwidth]{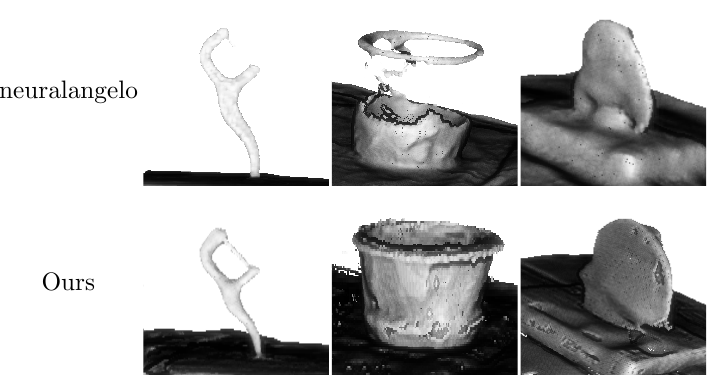}
\end{center}
    \caption{\textbf{Additional real-world comparisons with neuralangelo~\cite{neuralangelo}} Note that neuralangelo uses a different coordinate system and camera processing pipeline for COLMAP scenes, therefore the reconstructions are not perfectly aligned, but it can still be clearly seen that our method achieves better reconstruction quality on thin structures and translucent surfaces.}
    \label{fig:llff_neuralangelo}
\end{figure}

\ww{We show additional comparisons with neuralangelo~\cite{neuralangelo} in Fig~\ref{fig:llff_neuralangelo}. Note that as neuralangelo uses a different camera normalization for COLMAP scenes instead of Normalized Device Coordinate (NDC), which we use for our method and all other baselines, the reconstruction of neuralangelo is therefore not exactly aligned. We use an interactive viewer with Eye Dome Lighting~\cite{CloudCompare} and manually selected camera positions with close views for comparison. Regardless, it can be clearly seen that although neuralangelo excels at reconstructing smooth surfaces, it fails to faithfully reconstruct thin or translucent surfaces. Our method achieves a significant improvement over it in terms of thin and translucent surface reconstruction.}

\subsection{Ablation}

We should additional qualitative ablation of our method in Figure~\ref{fig:abla}. 
In addition, we show a comparison between the results after applying our TV surface regularization $\mathcal{L}_\delta$ and after applying the Eikonal constraint regularization used in most SDF optimization methods in Figure~\ref{fig:abla_ek}. Namely, in replace of TV surface regularization, we encourage the norm of the gradient of the surface field at every vertex to get close to $1$ via mean squared error:

\begin{align}
    \mathcal{L}_{ek} = \frac{1}{|\mathcal{V}|}  \sum_{\mathbf{x} \in \mathcal{V}} (||\nabla \hat{\delta}(\mathbf{x})||_2 - 1)^2 \,.
\end{align}

From Figure~\ref{fig:abla_ek}, it can be clearly seen that the Eikonal constraint is not sufficient to regularize and remove the noisy inner surfaces inherited from initialization. Moreover, it turns out to even harm the optimization by introducing additional surface floaters while trying to constrain the surface field into an SDF. This also shows that converting the surfaces extracted from a density field into proper SDF is a non-trivial task. 

In Figure~\ref{fig:abla_crop}, we show that normal regularization $\mathcal{L}_{\mathbf{n}_1}, \mathcal{L}_{\mathbf{n}_2}$ are insufficient for removing heavily biased surfaces initialized from Plenoxels, whereas $\mathcal{L}_\delta$ is more effective in this case.

\subsection{DTU Dataset}

\begin{table}
    \centering
\begin{tabular}{lrrrrr}
\toprule
                    {} &  37 &  40 &  63 &  69 &  110 \\
\midrule
Plen ($\sigma=10$)  &        1.90 &        1.86 &        1.86 &        2.04 &         1.96 \\
Plen ($\sigma=50$)  &        1.46 &        1.43 &        1.66 &        \second{}1.60 &         1.75 \\
Plen ($\sigma=100$) &        \second{}1.34 &        1.57 &        2.99 &        2.22 &         2.43 \\
NeuS                &        \first{}0.98 &        \first{}0.56 &        \second{}1.13 &        \first{}1.45 &         \second{}1.43 \\
Ours                &        \second{}1.34 &        \second{}1.36 &        \first{}0.99 &        1.91 &         \first{}1.37 \\
\bottomrule
\end{tabular}
\caption{\textbf{Chamfer distance $\downarrow$ on DTU scenes.} We color the \colorbox{first}{best} and \colorbox{second}{second best} surfaces.}
    \label{tab:dtu}
\end{table}

\begin{figure}[t]
\begin{center}
  \includegraphics[width=0.45\textwidth]{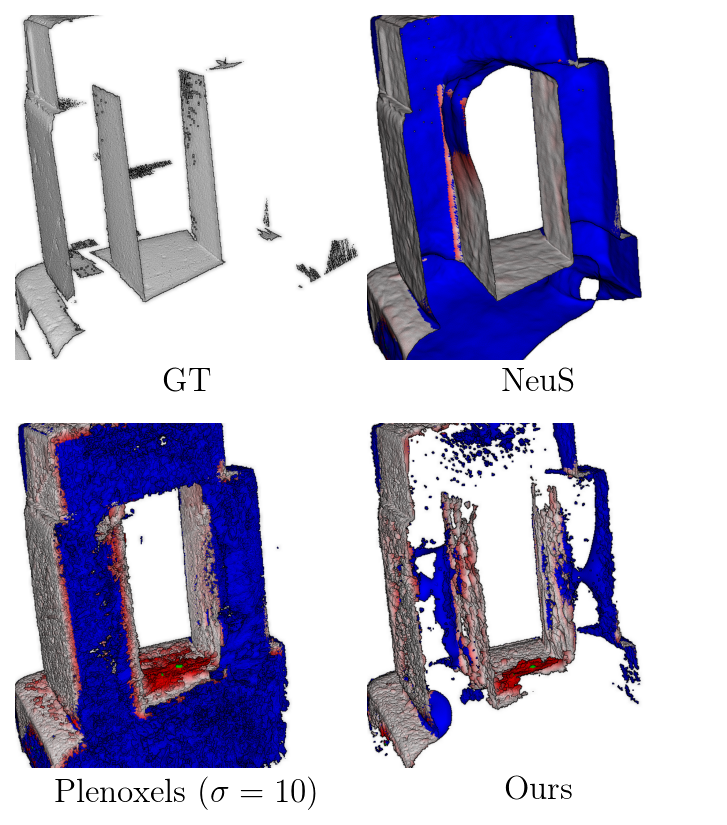}
\end{center}
   \caption{\textbf{Inside views of reconstructions on DTU dataset.} 
   Red color indicates the L1 error in reconstruction, and blue indicates the reconstruction masked out by the DTU official masks. Surfaces extracted from Plenoxels contain many noisy inner surfaces that had to be masked out during evaluation to achieve low Chamfer distance.
   }
\label{fig:dtu_crop}
\end{figure}

\begin{figure*}[t]
\begin{center}
  \includegraphics[width=0.95\textwidth]{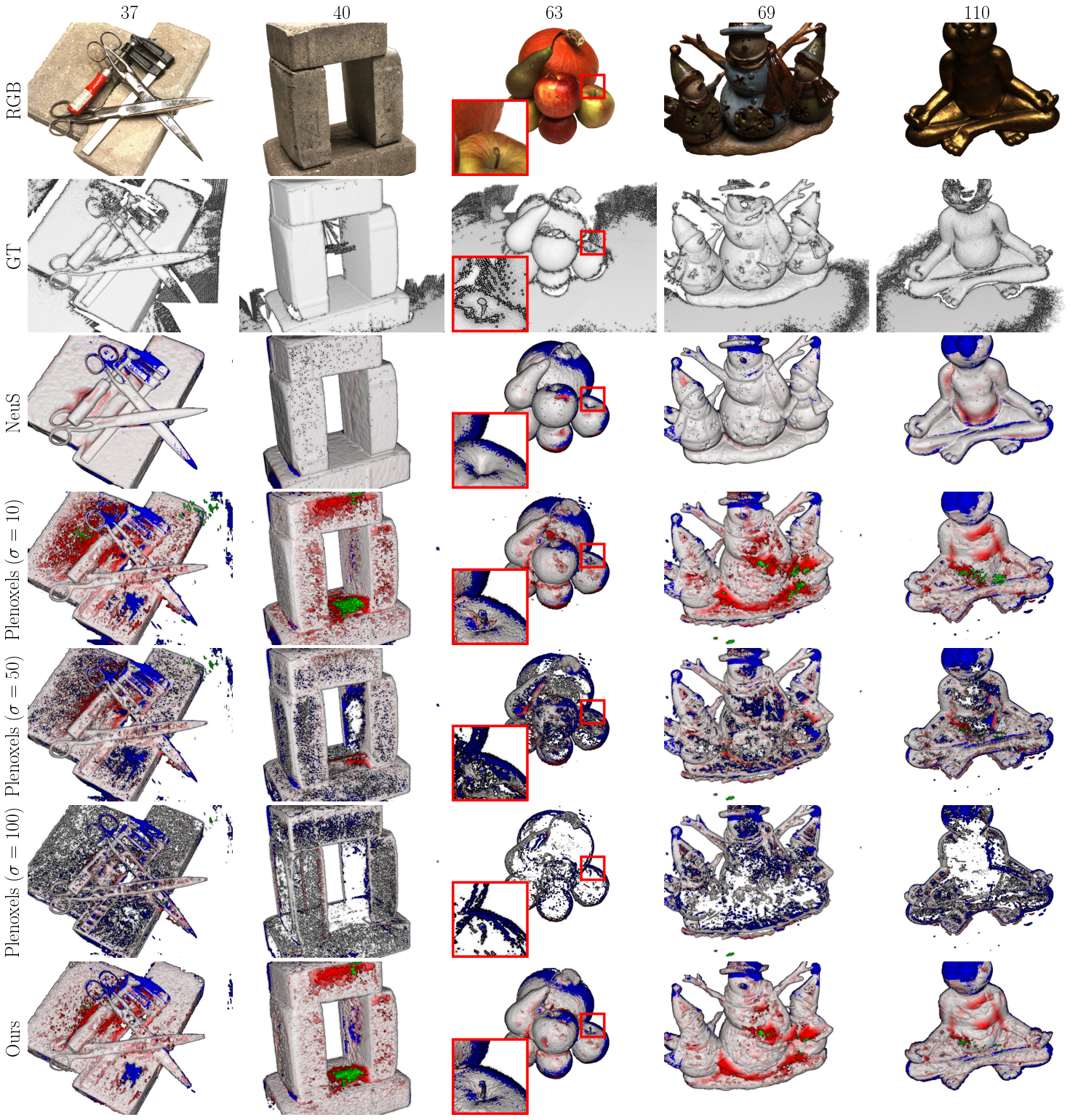}
\end{center}
   \caption{\textbf{Qualitative results on DTU \cite{dtu} dataset.} As the DTU scenes mainly contain smooth surfaces without any semi-transparent materials, our method does achieve state-of-the-art performance on this dataset. However, note that our method can still accurately capture the thin structure that is missed by NeuS in Scan 63. Moreover, our method can effectively correct the out-growing surface artifacts in Plenoxels. Red color indicates the L1 error in reconstruction, blue indicates the reconstruction masked out by the DTU official masks, and green indicates reconstructions that are too far away from reference and hence clipped during evaluation.
   }
\label{fig:dtu}
\end{figure*}

We additionally show reconstruction results on some DTU \cite{dtu} scenes in Figure~\ref{fig:dtu} and Table~\ref{tab:dtu}. We note that as DTU does not contain many thin structures or semi-transparent materials, but mostly smooth surfaces only, our method is therefore not expected to achieve state-of-the-art performance in this scenario. In fact, our method reconstructs reasonable surfaces, but performs worse than NeuS overall. This is mainly due to a lack of natural spatial smoothness constraint present in the MLP architecture of NeuS, which allows it to perform well on datasets like DTU that contain many smooth surfaces, but worse on our synthetic dataset with a focus on thin structures. 

We also note that although the qualitative comparison in Figure~\ref{fig:dtu} shows that our method can refine the level set surfaces extracted from Plenoxels by correcting the out-growing surfaces while preventing holes, the Chamfer distance does not always show an improvement. This is because the official DTU evaluation provides carefully created masks to remove reconstruction on parts that do not have proper reference geometry scanned by the depth scanner. This also excludes the majority of the inner surfaces from level set surfaces of Plenoxels, making their Chamfer distances much better; see Figure~\ref{fig:dtu_crop}.

\paragraph{Experiment Details} We compared with level set surfaces from Plenoxels \cite{plenoxels} and NeuS \cite{neus} trained with masks. We used the image masks provided by IDR \cite{idr} to set the background to white before training both Plenoxels and ours. For Plenoxels, we used the same hyperparameters for training on NeRF Synthetic dataset. We used slightly different hyperparameters from the ones we used for training NeRF Synthetic and Thin datasets. Namely, we modified the surface scalar learning rate to start with $10 ^ {-4}$ and end with $10^{-6}$. We increased $\lambda_\delta$ to $0.05$, $\lambda_\mathcal{H}$ to $10^{-3}$ and $\lambda_\alpha$ to $10^{-8}$. We also kept the truncated alpha compositing parameter $a$ at $5$ throughout training.



\end{document}